%% file: main.tex
\pdfoutput=1
\documentclass{banglab}

\input{commands}

\title{\texorpdfstring{\titleicon}{}\ourmethod{}: Self-Programming Hierarchical Memory for Language Agents}
\logos{udem,mila}
\newsavebox{\logoboxA}\newsavebox{\logoboxB}
\AtBeginDocument{%
  \sbox{\logoboxA}{\includegraphics[height=1pt]{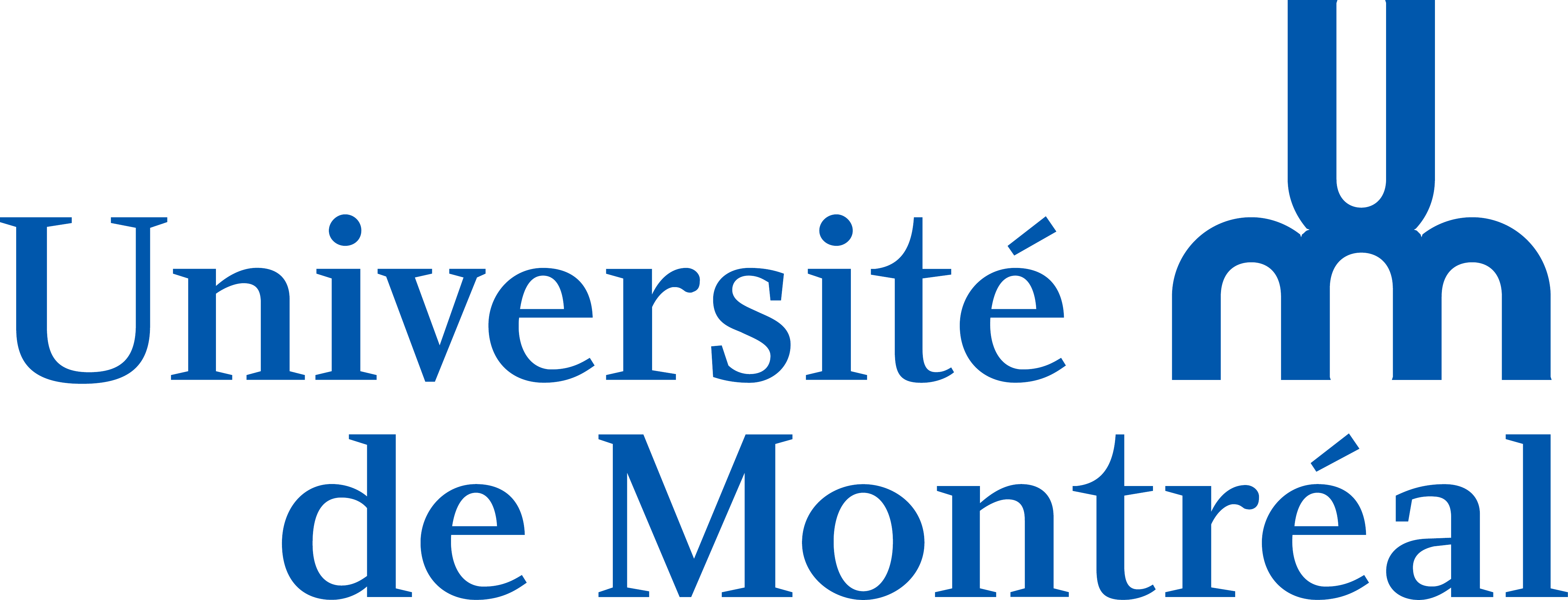}}%
  \sbox{\logoboxB}{\includegraphics[height=1pt]{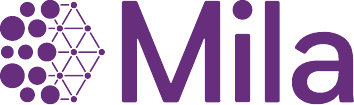}}}

\author[1,2]{Xiaoqiang Wang}
\author[1,2]{Bang Liu}

\affiliation[1]{Universit\'e de Montr\'eal}
\affiliation[2]{Mila -- Quebec AI Institute}

\abstract{%
Agent memory faces heterogeneous access needs: a single-hop question may require one piece of evidence, whereas a multi-hop question must combine evidence from multiple sources.
Predefined memory workflows cannot adapt to these varying needs.
Recent adaptive methods search or learn over memory components and their compositions, but the design space itself remains predefined.
We introduce \textbf{\ourmethod{}}, a self-evolving hierarchical memory system that organizes experience into executable memory programs for summaries, relational knowledge, reusable skills, and latent memory.
Open-ended program evolution searches the open design space of layer programs by rewriting how each layer is constructed, indexed, retrieved, and routed, thereby adapting both within-layer implementations and cross-layer composition.
At query time, reads traverse the hierarchy from coarse to fine and stop once sufficient evidence is found, descending to the original history when needed.
We further develop \arena{}, a unified runtime that places heterogeneous data and memory systems behind a common interface.
\ourmethod{} improves average task success by $10.1\%$ relative to the strongest adaptive-memory baseline, with $3.4\times$ fewer context tokens and $2.1\times$ faster inference.
}

\date{\today}
\correspondence{\email{bang.liu@umontreal.ca}}

\teaser{\input{figures/fig0_teaser.tex}}{Memory systems by memory organization and design search (left), and average test accuracy against latency per query (right).\label{fig:teaser}}

\begin{document}

\maketitle

\section{Introduction}
\label{sec:intro}

Language agents~\citep{yao2022react,wang2024survey,liu2025advances,ding2026combodied} increasingly act over long horizons in web navigation~\citep{zhou2023webarena}, software engineering~\citep{jimenez2023swe,hong2023metagpt} and computer use~\citep{xie2024osworld,wang2025oscar}, and self-evolve from experience~\citep{tao2024survey} through reflection on failures~\citep{shinn2023reflexion}, skill accumulation~\citep{wang2023voyager,zhao2024expel,ouyang2025reasoningbank,shi2026evolving} and code rewriting~\citep{zhang2025darwin}.
Across sessions, much of this experience survives only in memory.
Hand-designed pipelines organize it through rolling summaries in systems such as MemoryBank and MemGPT~\citep{zhong2024memorybank,packer2023memgpt}, facts and entity graphs~\citep{xu2025mem,rasmussen2025zep,edge2024graphrag}, skill stores~\citep{zhao2024expel,ouyang2025reasoningbank,wang2023voyager}, or memory written into parameters or latent state~\citep{wang2024memoryllm,behrouz2024titans,li2025memos,wang-etal-2025-r3mem}.
Yet these pipelines face \emph{access heterogeneity}, because the memory organization a query needs depends on the query, the record and the task~\citep{xu2026memgas,xu2026fluxmem,huang2026harnessmem}.
This heterogeneity appears across query types, since single-hop questions need raw turns~\citep{han2026rag}, multi-hop questions need cross-session links~\citep{gutierrez2024hipporag}, temporal questions need dated events~\citep{ge-etal-2025-tremu}, and repeated procedures need skills~\citep{wang2025agent}.
Adaptive computation in reasoning shows the same pattern, where dynamic shortcuts spend depth only on the steps that need it~\citep{wang2026system}.
A hand-designed pipeline hard-codes both which organizations exist and the rule that chooses among them, and a study of fourteen memory systems~\citep{guo2026agentnative} finds that no architecture dominates across tasks.

Adaptive memory addresses this rigidity in two main directions.
One direction searches a predefined design space by selecting and composing modules for encoding, storage, retrieval, and management~\citep{chen2026automem,xu2026fluxmem,zhang2025memevolve}.
The other, exemplified by Mem-alpha, learns policies for when and how to write, update, and retrieve memory~\citep{wang2025mem,wang2026mem,zhao2026vermem,yang2026cmimem}.
Despite their different forms, both directions must jointly determine \emph{how} experience is organized into memory and \emph{how} queries access it, since retrieval can only use structures created during writing.
As shown in \cref{fig:teaser}, this coupling places memory systems along two axes: memory organization and design search.
Among prior systems, only MemPro~\citep{liu2026mempro} opens the design itself to program search, but it evolves a single monolithic program over flat memory, without a hierarchy whose components can evolve and be accessed independently.

\begin{figure}[tp]
\centering
\includegraphics[width=\linewidth]{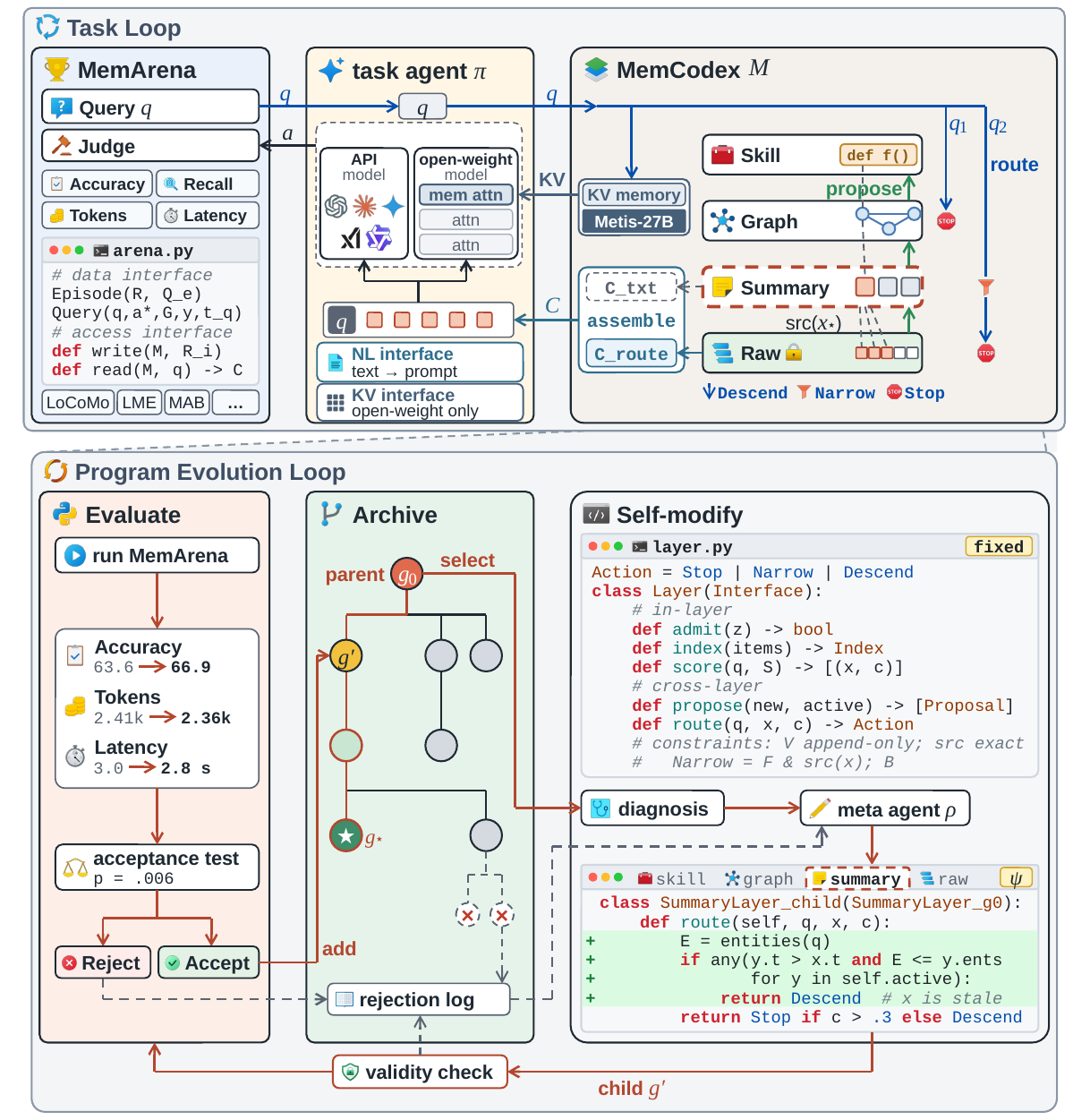}
\caption{\ourmethod{} workflow.
In the task loop (upper), \arena{} writes the record and poses each query $q$ to the task agent, which reads a context $C$ from the layer hierarchy, stopping at the coarsest sufficient layer or descending to source turns, and answers $a$.
In the program evolution loop (lower), run once per round, a parent is sampled from the archive of architectures $g=(\lambda,\psi)$, meta agent $\rho$ rewrites one method of one layer program $\psi_\lrung$, and validity checks and the acceptance test decide whether the child enters the archive.}\label{fig:plane}
\end{figure}

We therefore introduce \ourmethod{}, a hierarchical memory system whose layers are executable memory programs behind a fixed interface and runtime.
Program evolution adapts both the write and read code of individual layers and their cross-layer composition, that is, which layers are active and how items and reads pass between them.
Within this hierarchy, \ourmethod{} organizes \Lskill{}, \Lgraph{}, and \Lsum{} above an append-only raw layer, with \Lkv{} providing latent memory.

Inspired by the Darwin G\"odel Machine~\citep{zhang2025darwin} and HyperAgents~\citep{zhang2026hyperagents}, which evolve an agent by rewriting its own code, we apply program evolution to this memory hierarchy.
At each round, a meta agent draws a parent from the archive and rewrites one method of one layer or toggles one layer.
Restricting each child to one change makes the effect of that change attributable.
An acceptance test then keeps changes that improve accuracy or reduce token use without degrading the other, and accuracy-only gains must additionally pass a one-sided McNemar test~\citep{mcnemar1947note}.
Before this test, validity checks $\mathcal{K}$ reject changes that break the memory constraints.

The fixed interface exposes five methods in each layer, spanning both memory writing and reading.
On the write side, $\mathrm{admit}$ filters candidates, $\mathrm{index}$ keeps a bounded active set, and $\mathrm{propose}$ derives items for the layer above.
On the read side, $\mathrm{score}$ ranks candidates and $\mathrm{route}$ decides where the search proceeds next.
Reads move from coarse to fine layers, with $\mathrm{route}$ choosing \aStop{}, \aNarrow{}, or \aDescend{} until sufficient evidence is found or the search reaches the raw history.
Because every stored item retains its exact source turns, the original evidence remains recoverable even after lossy summarization.

To evaluate these mechanisms systematically across diverse sources, interaction formats, and evaluation protocols, we further develop \arena{}, a heterogeneous agent-source evaluation suite with a unified data interface, access interface, and metric suite covering benchmarks such as LoCoMo~\citep{maharana-etal-2024-evaluating}, LongMemEval (LME)~\citep{wu2025longmemeval}, and MemoryAgentBench (MAB)~\citep{hu2025evaluating}.
It separates the memory system from the task agent, as capability-level evaluation dissociates what a language model knows from how it reasons~\citep{wang-etal-2024-fac2e}.
Under one protocol, \arena{} spans seven sources, eleven baseline systems and two backbones, while each new source requires only one adapter.
Under this evaluation, \ourmethod{} gains $5.6$ average accuracy points over the strongest baseline with Qwen3.5-27B~\citep{qwen35_2026} and $3.6$ with the API model GPT-5.4-mini, while reading $3.4\times$ fewer tokens and answering $2.1\times$ faster.

\section{\ourmethod{}: Hierarchical Memory as Evolving Programs}\label{sec:method}

We model agent memory as a hierarchy of executable memory programs whose implementations and composition evolve with the task (\cref{fig:plane}).
A fixed five-method layer interface and runtime constrain how these programs interact while leaving their internal behavior open to evolution.

We model long-horizon memory as a task agent $\pi$ answering questions about a record $R$ that arrives as write chunks $R_1,\dots,R_n$ of timestamped turns $\leaf$, where each question \qry{} is posed once the record up to its question time has been written.
A memory system with architecture $g$ keeps a store \iset{} and serves $\pi$ through a write operation and a read operation:
\begin{equation}
  \iset\gets\texttt{write}_g(\iset,R_i),\qquad \ctx=\texttt{read}_g(\iset,\qry),\qquad a=\pi(\qry,\ctx).
  \label{eq:io}
\end{equation}
The write operation folds each chunk into the store, the read operation returns a context \ctx{} of at most \budget{} tokens, and $\pi$ answers from the question and that context.
To preserve exact evidence despite this bounded context, the store contains an append-only \emph{raw layer} \vset{} of all turns and lossy \emph{derived items} $x$ above it, each of which retains its exact \emph{source turns} $\mathrm{src}(x)\subseteq\vset$.

We represent each \emph{architecture} as $g=(\lambda,\psi)$ with two parts.
The layer programs $\psi$ implement the write and the read through five methods per layer.
Specifically, the write operation uses $\mathrm{admit}$, $\mathrm{index}$ and $\mathrm{propose}$, while the read operation uses $\mathrm{score}$ and $\mathrm{route}$.
The layer mask $\lambda$ independently switches derived layers on or off.
For the questions $\mathcal{Q}$ of a task, each query \qry{} has \emph{gold turns} $G(q)\subseteq\vset$.
The accuracy of $g$ is
$\mathrm{acc}(g)=\mathbb{E}_{\qry\sim\mathcal{Q}}\,\mathbf{1}[\pi(\qry,\ctx_g(\qry))\text{ correct}]$
and its cost is $\tokq(g)=\mathbb{E}_{\qry\sim\mathcal{Q}}|\ctx_g(\qry)|$.
Our objective is the most accurate architecture at no higher token cost than $g_0$.
We first define the layer programs $\psi$ and the read and write operations they implement.
We then define program evolution, which changes $g$ one method rewrite or one layer toggle at a time and keeps accepted architectures in an archive.

\subsection{Layer Programs \texorpdfstring{$\psi$}{psi}}\label{sec:formulation}

\paragraph{Layers.} Queries require different levels of abstraction, but every derived item must remain connected to the raw evidence from which it was produced.
We therefore order the four text layers $\lset=(\Lskill,\Lgraph,\Lsum,\Lverb)$ from coarse to fine, as in RAPTOR~\citep{sarthi2024raptor}, with store $\iset=\vset\cup\bigcup_{\lrung\neq\Lverb}\iset_\lrung$, where $\iset_\lrung$ holds the derived items of layer $\lrung$.
The \Lverb{} layer keeps every raw turn $\leaf\in\vset$ unchanged so that the original evidence remains available.
The \Lsum{} layer condenses each closed segment of raw turns into a compact account of that segment.
The \Lgraph{} layer stores timed assertions $x=(h,r,t,\tau)$ extracted from summaries, where entities $h$ and $t$ are linked by relation $r$ over the time $\tau$ for which the relation holds, so that a newer assertion can supersede an older one about the same entities.
The \Lskill{} layer stores executable procedures distilled from assertions that record the same procedure in several sessions.
We model a skill as a program rather than a text note because a procedure that runs can be checked against the sessions that demonstrate it.
Each derived item $\itemx$ is built from inputs $\srcof(\itemx)$ in the layer directly below, and its exact source turns $\mathrm{src}(\itemx)$ resolve recursively,
\begin{equation}
  \supp(\leaf)=\{\leaf\}\ \ (\leaf\in\vset),\qquad
  \supp(\itemx)=\textstyle\bigcup_{y\in\srcof(\itemx)}\supp(y)\ \ (\itemx\in\iset\setminus\vset).
  \label{eq:src}
\end{equation}
A read can therefore reach the raw turns underlying an assertion or a skill without traversing its derivation.
The set $\mathrm{src}(x)$ records the exact derivation of an item rather than a minimal evidence set.
In particular, a summary and its extracted assertions share the segment's turns, which \aNarrow{} and the raw-layer read subsequently refine.

\paragraph{Layer interface.} Local program evolution requires every method rewrite to remain within its target layer.
We therefore expose each text layer through five methods with distinct functionalities.
Within a layer, $\mathrm{admit}_\lrung$ filters proposals, while $\mathrm{index}_\lrung$ retains at most $K_\lrung$ items in the active index $\mathcal{I}_\lrung$.
Evicted items remain in the store.
The method $\mathrm{score}_\lrung(\qry,S)$ ranks the items $S$ allowed by the \emph{search scope} $\searchscope\subseteq\vset$ and returns the top candidate $x^\star$ with confidence \conf{}.
Across layers, $\mathrm{propose}_\lrung$ sends derived items upward, while $\mathrm{route}_\lrung(\qry,x^\star,\conf)$ returns \aStop{}, \aNarrow{} or \aDescend{}.
The runtime owns \texttt{assemble}, which packs \ctx{} under \budget{}.
In every architecture, turns are appended only to the raw layer and never changed.
Every derived item names the exact raw turns used to build it. \aNarrow{} can only restrict the search scope to the top candidate's source turns, $\searchscope\cap\mathrm{src}(x^\star)$.
Every context must also fit within \budget{} tokens.
These four \emph{memory constraints} ensure that any read can trace a derived item back to raw turns, however lossy the item is.

\begin{figure}[t]
\centering

\includegraphics[width=0.52\linewidth]{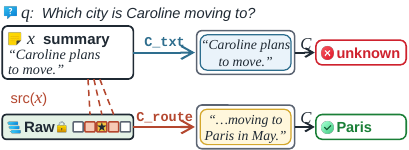}
\caption{Content and routing channels of \ourmethod{} for a question about Caroline. \chanTxt{} places summary $x$ in $C$, omitting Paris, while \chanRt{} follows $\mathrm{src}(x)$ to the raw turn naming Paris.}
\label{fig:channels}
\end{figure}
\paragraph{Content channel and routing channel.} A derived item can serve a query through two distinct channels, and the query-adaptive read uses the routing channel.
The channel is an architecture setting rather than a per-query choice, so only the read depth adapts to each query.
In the \emph{content channel} \chanTxt{}, the text of the top-scoring items enters \ctx{} in place of their evidence, as in summary, profile and skill memories~\citep{packer2023memgpt,chhikara2025mem0,zhao2024expel}.
In the \emph{routing channel} \chanRt{}, a retrieval hit leads to its source turns, as HippoRAG passes node probability to source passages~\citep{gutierrez2024hipporag}.
In the Caroline example of \cref{fig:channels}, the summary omits Paris, but routing through $\mathrm{src}(x)$ recovers the raw turn that names it.
\emph{Evidence preservation}, retaining every derived item's exact source turns, keeps this route available.

\paragraph{Query-adaptive read.} A fixed read depth can either skip useful abstractions or retrieve unnecessary detail.
The $\texttt{read}_g$ operation in \cref{eq:io} therefore visits layers from coarse to fine (\cref{alg:read}) and lets each $\mathrm{route}$ determine whether traversal should continue.
Only \aStop{} ends the read and adds the in-scope source turns $\mathrm{src}(x^\star)\cap\searchscope$ to \ctx{} in temporal order through the routing channel.
The \aNarrow{} action restricts the search scope to those source turns, while \aDescend{} preserves the current scope, and both actions continue to the next layer.
At the raw layer, the $k$ highest-scoring in-scope turns enter \ctx{}.
The routing channel therefore ensures that the task agent always answers from raw turns, even when the read stops at a skill whose source turns are the sessions that demonstrate the procedure.
Each query consequently incurs only the read depth it needs.

\paragraph{Write path.} Derived layers depend on lower-layer content, so the $\texttt{write}_g$ operation in \cref{eq:io} processes each appended chunk from bottom to top through $\mathrm{propose}$, $\mathrm{admit}$ and $\mathrm{index}$.
Each proposal to a layer combines the items newly admitted one layer below for the current chunk with the active items of that lower layer.
The runtime records every item read to form the proposal as an input.
At \Lskill{}, $\mathrm{admit}$ accepts a procedure only when replay on every demonstration session reproduces the recorded outcome.
The task agent's backbone writes the textual proposals.
In the initial architecture $g_0$, $\mathrm{index}$ retains items according to the access-and-recency heat of MemoryOS~\citep{kang2025memoryos}, $\mathrm{score}$ uses cosine similarity and $\mathrm{route}$ stops or narrows above per-layer thresholds.

\paragraph{Latent memory.} Some associations in the record have no representation in any text layer.
We feed each write chunk into Metis-27B~\citep{zhang2026metis}, a memory foundation model with stateful fast weights, and use the state of its memory blocks as \Lkv{}, a fixed-size key-value memory.
\Lkv{} is write-only, exposes no score or route, is built on the write path, and is not evolved.
Two interfaces provide memory to the task agent.
The natural language (NL) interface places the text context \ctx{} in the prompt and supports any agent.
The KV interface injects \Lkv{} into an open-weight agent's memory attention.
Because an API agent exposes neither weights nor activations, it uses only the NL interface and has no latent memory.

\begin{figure}[t]
\AlgPair{\textsc{QueryAdaptiveRead}}{alg:read}{%
  \Require query \qry, store \iset{} with an active index $\mathcal{I}_\lrung$ per layer,
           architecture $g$ with active layers $\lset_g$ ($\lambda=1$), budget \budget, channel $\chi\in\{\chanTxt,\chanRt\}$,
           and $\Fn{top}_k$ the first $k$ leaves by $\Fn{score}_{\Lverb}$
  \Ensure context \ctx
  \State $\searchscope\gets\vset$ \Comment{every raw turn}
  \For{$\lrung\in\lset_g\setminus\{\Lverb\}$, coarsest first}
    \State $S\gets\{\itemx\in\mathcal{I}_\lrung:\supp(\itemx)\cap\searchscope\neq\varnothing\}$
    \State $X\gets\Fn{score}_\lrung(\qry,S)$\label{line:score}
    \IfThen{$X=\varnothing$}{\textbf{continue}}
    \State $(\itembest,\conf)\gets X_1$ \Comment{top candidate}
    \BeginBox[AlgHi]
    \State $o\gets\Fn{route}_\lrung(\qry,\itembest,\conf)$\label{line:route} \Comment{searched method}
    \EndBox
    \If{$o=\aStop$}
      \IfThen{$\chi=\chanTxt$}{$Y\gets\Fn{text}(X_{1..10})$}
      \ElseThen{$Y\gets\supp(\itembest)\cap\searchscope$ in time order}
      \State \Return $\Fn{assemble}_\budget(Y)$
    \ElsIf{$o=\aNarrow$}
      \State $\searchscope\gets\searchscope\cap\supp(\itembest)$\label{line:narrow}
    \EndIf
  \EndFor
  \State \Return $\Fn{assemble}_\budget(\Fn{top}_k(\qry,\searchscope))$ \Comment{\aStop{} at \Lverb}
}{\textsc{ProgramEvolution}}{alg:evolve}{%
  \Require initial architecture $g_0$, evolve questions $Q$, meta agent $\rho$, validity checks
           $\mathcal{K}$, rounds $T$, parents $P$, children per parent $H$
  \Ensure archive $\mathcal{A}$ and the evolved architecture $g^{\star}$
  \State $\mathcal{A}\gets\{g_0\}$,\ \ $E_{g_0}\gets\Call{Evaluate}{g_0,Q}$
  \For{$r=1,\dots,T$}
    \ForAll{$n\sim w$, $j\le H$, \textbf{in parallel}}
      \State $(\lrung,u)\gets$ next target of $\mathcal{D}(g_n)$\label{line:target}
      \IfThen{$u=\mathrm{toggle}$}{$g'\gets g_n$, flip $\lambda_\lrung$}
      \BeginBox[AlgHi]
      \ElseThen{$g'\gets\rho(g_n,\mathcal{D}(g_n),\lrung,u,\mathcal{J}_n)$}\label{line:rewrite}
      \EndBox
      \If{some $\kappa\in\mathcal{K}$ fails on $g'$}\label{line:contract}
        \State $\mathcal{J}_n\gets\mathcal{J}_n\cup\{(g',\kappa)\}$;\ \textbf{continue}
      \EndIf
      \State $E_{g'}\gets\Call{Evaluate}{g',Q}$
      \IfThen{$\accept(g_n\to g')$}{add $g'$ under $n$}\label{line:cert}
      \ElseThen{$\mathcal{J}_n\gets\mathcal{J}_n\cup\{(g',E_{g'})\}$}
    \EndFor
  \EndFor
  \State \Return $\mathcal{A}$, $g^{\star}\gets\arg\max_{g\in\mathcal{A}}\mathrm{acc}_Q(g)$
}
\end{figure}

\subsection{Evolving the Architecture \texorpdfstring{$g$}{g}}\label{sec:evolve}

\paragraph{Architecture.} Joint evolution of layer behavior and composition requires each edit to remain local to one component.
The layer mask $\lambda\in\{0,1\}^{|\lset|-1}$ covers the derived layers, while the raw layer remains active.
Reads visit only the active layers.
Writes run through every layer up to the highest active one so that every proposal retains its type and derivation, including when an intermediate layer is switched off for reading.
The layer mask and cross-layer methods determine composition, while the within-layer methods determine implementation.
Starting from $g_0$, program evolution combines evaluator-guided program search~\citep{hu2025adas,romeraparedes2024funsearch,novikov2025alphaevolve}, the archive-based open-ended search of the Darwin G\"odel Machine~\citep{zhang2025darwin} and the meta agent of HyperAgents~\citep{zhang2026hyperagents}.
Meta agent $\rho$ runs on the task agent's backbone, so the agent programs its own memory.
\emph{Attributable edits} restrict every child to one method rewrite of one layer or one layer toggle, so each accepted change traces to one edit.
The acceptance test then accepts only improvements that do not trade accuracy against token cost.

\paragraph{Archive.} A single chain of edits cannot recover after reaching a dead end.
Following MemPro~\citep{liu2026mempro}, we therefore store accepted architectures in the archive $\mathcal{A}$, which forms a tree, while rejected children enter their parent's rejection log $\mathcal{J}_n$.
Each node stores evolve-question accuracy $\mathrm{acc}_Q$, tokens per question and verdicts $\mathbf{v}_g\in\{0,1\}^{|Q|}$.
A parent is drawn with probability proportional to
\begin{equation}
  w(n)=\sigma\!\Big(\tfrac{\mathrm{acc}_Q(g_n)-\overline{\mathrm{acc}}}{\hat\sigma}\Big)\cdot\frac{1}{1+|\mathrm{ch}(n)|}\cdot(1+d_n),
  \label{eq:parent}
\end{equation}
where $\sigma$ is logistic, $\overline{\mathrm{acc}}$ and $\hat\sigma$ are the archive accuracy mean and spread, $\mathrm{ch}(n)$ the children of $n$ and $d_n$ its normalized crowding distance on the accuracy-cost Pareto front (zero off it), so the weight favors accurate, rarely expanded nodes on the front.
Any archived architecture can become a parent, and a child may contain any code that satisfies the layer interface.
Program evolution therefore remains open-ended rather than following a single chain of edits.

\paragraph{Self-modification.} Each local edit should address an observed failure.
The parent's diagnosis $\mathcal{D}(g)$ therefore examines failed traces and refused or evicted items, then assigns each failure to the first method that loses a gold turn.
Children target successive methods or layer toggles according to fault count, so siblings make distinct edits.
For a method rewrite, meta agent $\rho$ reads the target layer's code, diagnosis and rejection log, then returns a subclass that overrides only that method (\cref{alg:evolve}).
Validity checks $\mathcal{K}$ reject any child that violates a memory constraint on fixture queries.

\paragraph{Acceptance test.} Validity checks determine whether a child preserves the memory constraints, but they do not determine whether it trades accuracy for token cost.
The acceptance test therefore compares child and parent on the same evolve questions with the task agent and judge frozen, accepting the child exactly when
\begin{equation}
  \accept(g\to g')\iff
  \Delta_{\mathrm{tok}}\le0\ \wedge\ \Delta_{\mathrm{acc}}\ge0\ \wedge\
  \big[(\Delta_{\mathrm{acc}}>0\ \wedge\ p<\alpha)\ \vee\ \Delta_{\mathrm{tok}}<0\big],
  \label{eq:cert}
\end{equation}
where $\Delta_{\mathrm{acc}}$ and $\Delta_{\mathrm{tok}}$ are child-minus-parent changes in accuracy and tokens per question, $p$ is the one-sided exact McNemar $p$-value~\citep{mcnemar1947note} on discordant pairs of $\mathbf{v}_g$ and $\mathbf{v}_{g'}$, and $\alpha=0.05$.
The evolved architecture $g^\star$ is the archived architecture with the highest evolve-split accuracy.
\section{\arena{}: A Heterogeneous Agent-Source Evaluation Suite}\label{sec:arena}
\begin{figure}[t]
\centering
\includegraphics[width=\textwidth]{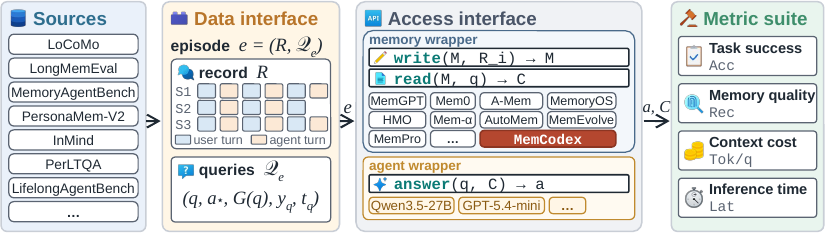}
\caption{The \arena{} evaluation suite, with heterogeneous sources behind unified data and access interfaces and a metric suite.}
\label{fig:arena}
\end{figure}
Memory systems are reported on sources with different evidence formats, time representations and access protocols, so their results are difficult to compare directly.
\arena{} addresses this problem by placing heterogeneous sources and memory systems behind one data interface, one access interface and one metric suite, which allows any memory module to be evaluated with any task agent under a fixed backbone and judge.
It currently spans seven sources, eleven baseline systems and two backbones.
We release the suite together with its adapters and wrappers.

\paragraph{Data interface.} Source adapters map corpora with different evidence and time representations into an episode $e=(R,\mathcal{Q}_e)$ (\cref{fig:arena}).
The record $R$ orders sessions of timestamped, speaker-tagged turns, while each tuple $(\qry,a^\star,G(q),y_q,t_q)\in\mathcal{Q}_e$ specifies a question, reference answer, gold turns, question type and question time, posed after the record through $t_q$ is written.
Source adapters obtain $G(q)$ from evidence annotations for LoCoMo and LongMemEval and from task construction for \pbench{} and MAB. They cover seven sources~\citep{maharana-etal-2024-evaluating,wu2025longmemeval,hu2025evaluating,jiang2025personamem,li-etal-2025-inmind,du-etal-2024-perltqa,zheng2025lifelongagentbench}, and adding a source requires only a source adapter.
\pbench{}, our fourth benchmark, poses five preference or procedure questions for each of 300 PersonaMem-V2 and InMind records, yielding 1{,}500 questions.

\paragraph{Access interface.} To separate memory quality from task-agent quality, \arena{} wraps memory systems and task agents independently.
A memory wrapper exposes $\texttt{write}(\iset,R_i)\to\iset'$ and $\texttt{read}(\iset,\qry)\to C$, where $C$ may carry \Lkv{}.
An agent wrapper separately exposes $\texttt{answer}(\qry,C)\to a$ for task agent $\pi$, using the NL interface for text and, for open-weight agents, the KV interface for \Lkv{}.
The harness writes each record into a fresh store in 20-turn chunks and poses each question after the last chunk written before $t_q$, so stores are never shared across records or users and only $g$ is shared.
This separation allows any task agent to be evaluated with any memory module.
Within each comparison, the harness fixes the backbone and judge, so systems differ only in memory.

\paragraph{Metric suite.} \arena{} measures both task success and memory cost.
Accuracy $\mathrm{acc}$ is the judge verdict on $a$ against $a^\star$ under the Mem0 rubric~\citep{chhikara2025mem0}.
Where gold turns are annotated, evidence recall is $\mathrm{Rec}=\mathbb{E}_{\qry}[|G(q)\cap C|/|G(q)|]$.
Context cost \tokq{} counts prompt tokens per question, and inference time measures wall-clock latency from \texttt{read} to the last answer token.

\section{Experiments}\label{sec:experiments}

\paragraph{Benchmarks and splits.} We evaluate every system through \arena{} on LoCoMo~\citep{maharana-etal-2024-evaluating}, LongMemEval (S split)~\citep{wu2025longmemeval}, MemoryAgentBench (MAB)~\citep{hu2025evaluating} and \pbench{}.
Following MemEvolve~\citep{zhang2025memevolve} and MemPro~\citep{liu2026mempro}, which evolve on a held-out slice of each benchmark, we take 20 percent of each benchmark's questions as the evolve split $Q$.
We stratify $Q$ by question category, such as single-hop, multi-hop, temporal and open-domain questions on LoCoMo, the five abilities of LongMemEval, the four competencies of MAB, and preference against procedure questions on \pbench{}.
We never read the test split during search.

\paragraph{Baselines.} Unstructured baselines are 32k-token full context and dense RAG with BGE~\citep{xiao2024cpack} over 512-token chunks under $\budget=4{,}096$.
Hand-designed baselines are MemGPT~\citep{packer2023memgpt}, A-Mem~\citep{xu2025mem}, Mem0~\citep{chhikara2025mem0}, MemoryOS~\citep{kang2025memoryos} and HMO~\citep{liu2026hmo}.
Adaptive or evolved baselines are Mem-$\alpha$~\citep{wang2025mem}, AutoMem~\citep{chen2026automem}, MemEvolve~\citep{zhang2025memevolve} and MemPro~\citep{liu2026mempro}.
Search-based baselines share the evolve split, meta agent and 64-child budget.

\paragraph{Models and search.} We evaluate the open-weight Qwen3.5-27B~\citep{qwen35_2026} and the API model GPT-5.4-mini.
Per run, one backbone serves every system as task agent $\pi$, writes derived items, and serves as meta agent $\rho$ at temperature 0.2 with one call per child.
Qwen3.5-27B reads text through the NL interface and Metis-27B~\citep{zhang2026metis} state \Lkv{} through the KV interface, while GPT-5.4-mini uses only the NL interface.
Specifically, Qwen3.5-27B takes \Lkv{} from Metis-27B, an open-weight memory foundation model that inserts fast-weight memory blocks into a frozen Qwen3.5-27B and updates them by test-time training in the forward pass with a gated delta rule, as in Titans~\citep{behrouz2024titans}.
For each record, we run Metis-27B, extract its memory blocks' key-value state, and inject it through the KV interface.
Because the two models share the frozen backbone, injection is native and requires no alignment.
Without weight access, GPT-5.4-mini has no latent memory.
GPT-4o-mini~\citep{hurst2024gpt} is the judge.
Evolution uses $T=8$ rounds, $P=4$ parents and $H=2$ children per parent.
Search-based results average three seeds.

\begin{table}[t]
\caption{Quantitative results of \ourmethod{} with Qwen3.5-27B and GPT-5.4-mini as task agents.
\textbf{Bold} marks the best per column, \underline{underline} marks the strongest baseline, and parentheses give \ourmethod{}'s margin over it.
Tok/q counts thousands of prompt tokens per question, and LME, MAB and PMind abbreviate LongMemEval, MemoryAgentBench and \pbench{}.}
\label{tab:main}
\centering
\footnotesize
\setlength{\tabcolsep}{2.6pt}
\input{tables/main}
\end{table}

\subsection{Main Results}\label{sec:main}

\textbf{\ourmethod{} is the most accurate system on all four benchmarks and reads the fewest tokens.} It averages $60.8$ percent, $5.6$ points above MemPro and $7.2$ above full context, while reading $2.0$k tokens per question against $6.8$k and $21.4$k (\cref{tab:main}).
It answers in $2.7$ seconds, $2.1\times$ faster than MemPro and $3.6\times$ faster than full context (\cref{fig:teaser}, right).
Seed standard deviations are at most $0.8$ points.
Margins over MemPro peak on \pbench{} ($+6.9$) and LoCoMo ($+5.9$), which we attribute to latent memory carrying recurring preferences and routes linking cross-session evidence (\cref{fig:results}, middle).
The margin is smallest on LongMemEval ($+4.5$), whose records already isolate evidence.

\textbf{Evidence preservation puts $g_0$ ahead, and evolution supplies most of the margin.} The initial architecture $g_0$ averages $56.8$ percent at $2.4$k tokens, $1.6$ points above MemPro, which we attribute to derived items routing reads to their exact source turns.
Program evolution adds $4.0$ points while reducing tokens by 17 percent.
LoCoMo evidence recall rises from MemPro's $79.3$ percent to $88.4$ percent (\cref{fig:results}, left).
With GPT-5.4-mini, which evolves its own architecture through the NL interface alone, \ourmethod{} leads MemPro by $3.6$ points, matching the margin of the Qwen3.5-27B variant without \Lkv{}, which averages $58.8$ and leads MemPro by $3.6$, so latent memory contributes $2.0$ points under Qwen3.5-27B.

\begin{table}[t]
\caption{Ablations of \ourmethod{} on hierarchy depth, evidence channel and evolution, with one change per row under the same search budget.
Avg. averages the four benchmarks, Rec. denotes LoCoMo evidence recall, and parentheses give the change from \ourmethod{}.}
\label{tab:ablation}
\centering
\footnotesize
\setlength{\tabcolsep}{1.4pt}
\input{tables/ablation}
\end{table}

\subsection{Ablation}\label{sec:ablation}

\Cref{tab:ablation} tests the two claims behind \ourmethod{}, the hierarchy with routed evidence and per-method evolution, using variants evolved under the same budget.
To test hierarchy depth, we fix the number of layers to one (the raw layer), two (adding \Lsum{}), three (adding \Lgraph{}), or four (the full hierarchy with \Lskill{}).
To isolate the evidence channel, variants place derived-item text in \ctx{} instead of source turns, or alongside them.
The evolution variants retain four layers but change how their programs are obtained.
No evolution keeps $g_0$.
Module selection chooses each method among predefined implementations as AutoMem does.
Layer rewrites change one layer program per child as the slot-wise evolver of MemEvolve does, while program rewrites change every layer per child as MemPro does.
Finally, no acceptance test archives every child that passes the validity checks, and no archive follows one greedy chain.
Read policies are compared under RQ5, and the variant without \Lkv{} appears with the main results.

\paragraph{RQ1: Does the hierarchy matter?}
\textbf{Each added layer improves accuracy while context remains near $2$k tokens.} One, two, three, and four layers score $54.1$ ($-6.7$), $56.5$ ($-4.3$), $59.0$ ($-1.8$), and $60.8$, while Tok/q remains between $1.8$k and $2.3$k.
Because the context cost stays flat, we attribute the gains to reads that stop at the right level rather than to more context.
\Lgraph{} contributes the largest gain, $+2.5$ points, which we attribute to its cross-session links.

\textbf{Routing to source turns preserves evidence more efficiently than reading derived text.} Against full \ourmethod{} at $60.8$ accuracy, $88.4$ Rec.\ and $2.0$k Tok/q, the content channel reaches $55.4$ ($-5.4$) with $63.0$ Rec.\ and $3.9$k Tok/q.
Content plus routing restores Rec.\ to $88.7$ but reaches only $59.6$ ($-1.2$) at $4.6$k Tok/q.
Derived text therefore more than doubles context when added beside source turns and still dilutes the evidence used for answering.

\paragraph{RQ2: Does open-ended per-method evolution beat the prior adaptive lines?}
\textbf{Per-method evolution outperforms matched variants of the three prior search lines on the same four-layer hierarchy.} \ourmethod{} scores $60.8$, compared with $56.8$ ($-4.0$) for $g_0$, $58.3$ ($-2.5$) for module selection, $57.2$ ($-3.6$) for layer rewrites, and $56.7$ ($-4.1$) for program rewrites.
Menus remain limited to implementations anticipated by the designer.
Bundled rewrites remain near $g_0$, which we attribute to several changes being harder to preserve and attribute at once, with validity checks rejecting 41 of 64 layer-rewrite children against 9 of 64 per-method children.

\textbf{The acceptance test prevents evolve-split overfitting, and the archive beats a single greedy chain.} Without the acceptance test, accuracy falls to $58.1$ ($-2.7$) because noisy children enter the archive.
Without the archive, a greedy chain reaches $59.2$ ($-1.6$) because it cannot return to a useful ancestor after a dead end.

\begin{figure}[t]
\centering
\input{figures/fig4_dynamics.tex}
\caption{Evolution dynamics of \ourmethod{} across three search seeds, as gain over round 0 on LoCoMo for the search-based systems (left), the fraction of test reads stopping at each layer under $g^\star$ by LoCoMo question type and by benchmark (middle), and LoCoMo test accuracy by question type under always-raw, stop-at-first-hit and adaptive reads (right).
LME and PMind abbreviate LongMemEval and \pbench{}.}
\label{fig:dynamics}
\end{figure}
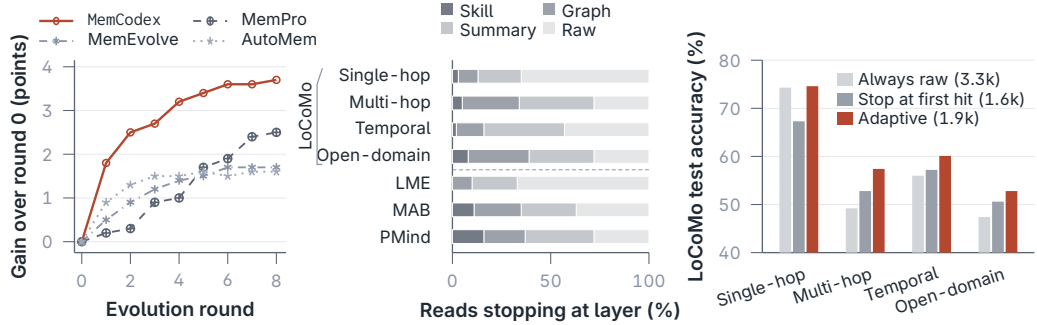

\subsection{Evolution Dynamics}\label{sec:dynamics}

\begin{table}[t]
\centering
\small
\setlength{\tabcolsep}{2.2pt}
\caption{Accepted edits along the chain from $g_0$ to $g^\star$ on LoCoMo, median seed, with the round of acceptance, the rewritten method, evolve-split and test accuracy, and thousands of prompt tokens per test question.}
\label{tab:chain}
\input{tables/chain}
\end{table}

\paragraph{Setting.} These analyses use Qwen3.5-27B and three search seeds per benchmark, each with $T=8$ rounds of eight children, or 64 children per search.
Per-round test accuracy is computed after the search from archived checkpoints of the best evolve-split architecture and averaged over seeds.
Retention divides the test-split gain of $g^\star$ over $g_0$ by its evolve-split gain.
Transfer evaluates one benchmark's $g^\star$ unchanged on every test split.

\paragraph{RQ3: How does accuracy change across rounds?} \textbf{\ourmethod{} gains more and keeps gaining, while the three search-based baselines gain less and flatten.} \ourmethod{} gains $3.7$ points, from $63.4$ to $67.1$, including $1.8$ in the first round from a route rewrite.
MemPro gains $2.5$ ($58.7$ to $61.2$) through whole-program rewrites, mostly after round four.
MemEvolve gains $1.7$ ($58.9$ to $60.6$) and flattens after round six.
AutoMem gains $1.6$ ($58.5$ to $60.1$) and plateaus from round three.
Per-method evolution keeps finding acceptable edits in later rounds because each child changes one method, so a small gain is attributable and passes the test.

\Cref{tab:chain} traces the accepted edits from $g_0$ to $g^\star$ on LoCoMo.
The first edit rewrites the $\mathrm{route}$ method of \Lsum{} so that a stale summary cannot stop the read, raising evolve-split accuracy from $63.6$ to $66.9$.
Later edits each add $0.2$ to $1.3$ evolve-split points, while Tok/q falls with every accepted edit from $2.31$k to $1.93$k.
The round-5 child descends from the round-3 node rather than the round-4 leader, because the archive retains every accepted architecture as a parent.
Every accepted edit names one method of one layer.

\begin{table}[t]
\caption{Test accuracy under record-disjoint splits with Qwen3.5-27B, mean of three seeds, with the best per column in \best{bold} and the second \second{underlined}, and the change of the average against \cref{tab:main} in parentheses.}
\label{tab:recsplit}
\centering
\small
\setlength{\tabcolsep}{4pt}
\input{tables/record_split}
\end{table}

\paragraph{RQ4: Do accepted gains hold on the test split?} \textbf{The acceptance test rejects improvements that primarily fit evolve-split noise.}
With the acceptance test, the test split retains 53 to 78 percent of the evolve-split gain, including $3.7$ of $5.8$ points on LoCoMo.
By contrast, accepting every child that passes $\mathcal{K}$ admits 53 to 56 of 64 children and ends 1.3 to 2.0 points higher on the evolve split, but retains only 10 to 24 percent of its gain.
Retention is lowest on LongMemEval, which has 100 evolve questions.
Under record-disjoint splits that hold out whole records (\cref{tab:recsplit}), the margin over MemPro is $5.0$ points and the evolution gain is $3.4$, so at most $0.6$ points are record-specific.
The held-out records are 2 of the 10 LoCoMo conversations, 29 of the 146 MAB records and 60 of the 300 \pbench{} records, while LongMemEval already pairs each question with its own record.
The evolve split takes 20 percent of the questions of the evolve records and the test split every question of the held-out records, so no question of a held-out record is seen at any stage.
Every baseline and $g_0$ lose $0.4$ points against \cref{tab:main}, whereas \ourmethod{} loses $1.0$.

\paragraph{RQ5: Does the read adapt to the question and the record?} \textbf{Adaptive reads match the best fixed depth on every question type at a fraction of the tokens.} No fixed policy wins everywhere.
Always raw leads on single-hop questions ($74.3$), and stop at first hit on the other three.
The adaptive read scores $74.6$, $57.4$, $60.1$ and $52.8$ on single-hop, multi-hop, temporal and open-domain questions, matching or exceeding the best fixed policy on every type and leading by $4.6$ points on multi-hop, while reading $1.9$k tokens against $3.3$k for always raw (\cref{fig:dynamics}, right).
Under $g^\star$, 65 percent of single-hop reads reach the raw layer, whereas 72 percent of multi-hop reads stop above it.
Across ten LoCoMo conversations, 37 to 61 percent of all reads and 58 to 83 percent of multi-hop reads stop above raw, so depth also varies with the record.

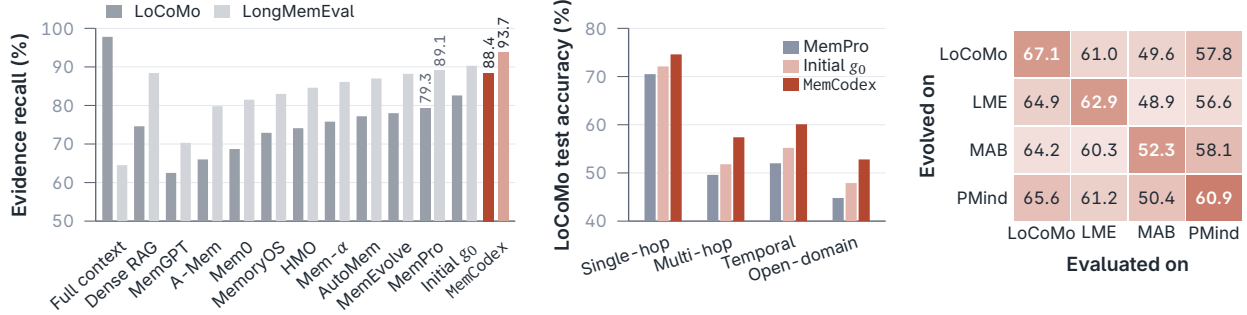
\begin{figure}[t]
\centering
\input{figures/fig6_results}
\caption{Evidence recall, question-type accuracy and cross-benchmark transfer of \ourmethod{}, as evidence recall on LoCoMo and LongMemEval for every system of \cref{tab:main} (left), LoCoMo test accuracy by question type for MemPro, the initial architecture $g_0$ and the evolved architecture (middle), and test accuracy of each $g^\star$ on each test split, darker for larger gains over $g_0$ and bold on the diagonal (right).}
\label{fig:results}
\end{figure}

\paragraph{RQ6: Are evolved architectures task-specific?} \textbf{Each task evolves a distinct architecture, although useful structure transfers.}
Every search rewrites a route method, and three searches also rewrite a write-path method.
Each native $g^\star$ beats the best transferred architecture by 1.5 to 2.8 points, while transferred architectures retain 0.6 to 2.4 points over $g_0$, which is about one third of the native gain (\cref{fig:results}, right).
The LongMemEval architecture switches \Lskill{} off and adds only $0.6$ on MAB and $0.9$ on \pbench{}, where 11 and 16 percent of reads stop at \Lskill{}.

\section{Related Work}\label{sec:related}

\paragraph{Agent memory evolution.} Agent memory has moved from hand-designed stores toward structures and operations that adapt to the task~\citep{hu2025memory,wu2026memoryera}.
The first line of work hand-designs architectures as paged tiers, fact and entity stores, hierarchical or graph indexes, and latent or parametric memory~\citep{packer2023memgpt,zhong2024memorybank,chhikara2025mem0,xu2025mem,rasmussen2025zep,kang2025memoryos,liu2026hmo,sarthi2024raptor,gutierrez2024hipporag,wu2026gam,chen2026memforest,hsu2026organize,wang2024memoryllm,behrouz2024titans,li2025memos,zhang2026metis,wang-etal-2025-r3mem}.
MemGPT~\citep{packer2023memgpt}, for example, pages information between the context window and external storage following an operating-system model.
Beyond hand design, a second line instead learns policies that decide what to write, update, consolidate, forget, or retrieve~\citep{wang2025mem,zhao2026vermem,yang2026cmimem,huang2026memcon,li2026memskill,zhou2026duallayer,chen2026adamem,lin2026memma,ye2026coevomem,wang2026mem}.
Mem-$\alpha$~\citep{wang2025mem}, for instance, trains its write and update operations with reinforcement learning.
More recently, memory design itself has become a search problem~\citep{chen2026automem,xu2026fluxmem,zhang2025memevolve,xiong2026alma,liu2026evolvemem}.
Closest to our setting, MemPro~\citep{liu2026mempro} evolves the entire construction and retrieval pipeline as one program over a version tree of runnable implementations.
By contrast, \ourmethod{} evolves the layer programs and hierarchy composition one method at a time, while every derived item keeps its exact source turns, so evidence remains recoverable throughout evolution.

\paragraph{Program search.} More broadly, program search lets agents improve their own code or the harness around a model under empirical evaluation.
One line automates agent and harness design~\citep{hu2025adas,lee2026metaharness,lin2026harness,zhang2026agenticevo,wu2026modularrsi}.
ADAS~\citep{hu2025adas}, for instance, lets a meta agent program new agents and archive its discoveries.
Another line extends this approach to self-referential agents that rewrite their own code~\citep{schmidhuber2007godel} and retain variants through empirical validation rather than the proof demanded by the G\"odel machine~\citep{zhang2025darwin,zhang2026hyperagents,iacob2026redqueen,liu2026mendel,srikanth2026rsi}, while acceptance rules for such changes are studied directly~\citep{shawn2026pace}.
The Darwin G\"odel Machine~\citep{zhang2025darwin} exemplifies this line by growing an archive of self-modifying coding agents validated on benchmarks.
HyperAgents~\citep{zhang2026hyperagents} further merge the task agent and meta agent into one editable program.
By contrast, in \ourmethod{} the search unit is one method of one memory layer behind a fixed interface, and a child enters the archive only after validity checks and an acceptance test.

\section{Conclusion}\label{sec:conclusion}

We propose \ourmethod{}, a self-evolving hierarchical memory system that organizes experience into executable layer programs for summaries, relations, skills and latent memory.
It evolves each layer through attributable per-method edits behind fixed interfaces and keeps every derived item's exact source turns.
Reads traverse the hierarchy from coarse to fine and descend to source turns as needed.
Through \arena{}, it outperforms hand-designed and adaptive memory systems across benchmarks and backbones at lower cost.
The experiments credit evidence preservation with the initial lead and per-method evolution with most of the rest, while the acceptance test carries evolve-split gains to the test split and evolved architectures remain task-specific yet transfer in part.
Next, we will evolve the runtime and the layer set themselves under record-disjoint deployment.

\bibliographystyle{plainnat}
\bibliography{custom,claudetodo,anthology}

\end{document}

%% file: commands.tex
\usepackage{multirow}
\usepackage{pgfplots}
\pgfplotsset{compat=1.18}
\usepackage{nicefrac}
\usepackage{array}
\usepackage{makecell}
\usepackage{tabularx}
\usepackage{comment}
\usepackage{pifont}
\usetikzlibrary{arrows.meta,positioning,fit,backgrounds,calc,decorations.pathreplacing,patterns,shapes.geometric,shapes.symbols,shadows.blur,matrix}

\usepackage{algorithm}                 
\usepackage[noEnd=true,indLines=true,italicComments=false,
            commentColor=McGrey]{algpseudocodex}

\theoremstyle{plain}

\theoremstyle{definition}

\theoremstyle{remark}

\definecolor{dkgreen}{rgb}{0,0.6,0}
\definecolor{gray}{rgb}{0.5,0.5,0.5}
\definecolor{mauve}{rgb}{0.58,0,0.82}

\definecolor{dgreen}{rgb}{0.412,0.741,0.271}
\definecolor{dblue}{rgb}{0.220,0.325,0.639}
\definecolor{dred}{rgb}{0.933,0.122,0.137}

\definecolor{memhighlight}{RGB}{227,242,253}
\definecolor{memframe}{RGB}{21,101,192}
\definecolor{abstainhighlight}{RGB}{255,243,224}
\definecolor{abstainframe}{RGB}{230,81,0}

\definecolor{bestcell}{RGB}{232,245,233}    
\definecolor{rowshade}{RGB}{248,248,248}    
\definecolor{csframe}{RGB}{46,125,50}       
\definecolor{csbg}{RGB}{250,253,250}        
\definecolor{memcolor}{RGB}{25,118,210}     
\definecolor{abstaincolor}{RGB}{239,108,0}  
\definecolor{darkgreen}{RGB}{27,94,32}
\definecolor{wacell}{RGB}{232,240,254}      
\definecolor{labcell}{RGB}{255,243,224}     

\definecolor{g1}{HTML}{b3e2cd}
\definecolor{r1}{HTML}{fdcdac}
\definecolor{w1}{HTML}{cbd5e8}
\definecolor{b1}{HTML}{fff7bc}

\definecolor{lr}{HTML}{bebada}
\definecolor{fr}{HTML}{fccde5}

\definecolor{Lavender}{HTML}{BF94E4}

\definecolor{l1}{RGB}{189,215,238}
\definecolor{l2}{RGB}{222,235,247}
\definecolor{l3}{RGB}{255,230,153}
\definecolor{l4}{RGB}{248,203,173}
\definecolor{l5}{RGB}{244,177,131}

\newtcbox{\columntcbox}{highlight math style={
        colback=gray!30,
        arc=2pt,
        outer arc=2pt,
        boxrule=0pt,
        top=2pt,
        bottom=2pt,
        left=2pt,
        right=2pt,
    }
}

\colorlet{LightLavender}{Lavender!35!}
\newtcbox{\inlinetcbox}[1][]{on line, 
        boxsep=2pt, left=0pt,right=0pt,top=0pt,bottom=0pt,
        colframe=white,colback=LightLavender,  
        highlight math style={enhanced}, #1
}

\newcolumntype{C}[1]{>{\centering}m{#1}}

\makeatletter
\newcommand{\xMapsto}[2][]{\ext@arrow 0599{\Mapstofill@}{#1}{#2}}
\def\Mapstofill@{\arrowfill@{\Mapstochar\Relbar}\Relbar\Rightarrow}
\makeatother

\definecolor{McVerb}{HTML}{1d4ed8}   
\definecolor{McCoarse}{HTML}{ea580c} 
\definecolor{McSkill}{HTML}{16a34a}  
\definecolor{McParam}{HTML}{7c3aed}  
\definecolor{McSoft}{HTML}{eff6ff}   
\definecolor{McWarn}{HTML}{dc2626}   
\definecolor{McGrey}{HTML}{475569}
\definecolor{McInk}{HTML}{22303C}    
\definecolor{McMute}{HTML}{8A94A6}   
\definecolor{PsContext}{HTML}{DCE7F2}
\definecolor{PsNotes}{HTML}{E6E1F1}
\definecolor{PsGraph}{HTML}{DDEFE3}
\definecolor{PsSkill}{HTML}{FAEAD5}
\definecolor{PsLayer}{HTML}{F6DDE3}
\definecolor{PsEdge}{HTML}{9AA3B0}    
\definecolor{PsFrame}{HTML}{6F7885}   

\newcommand{\ourmethodname}{MemCodex}
\newcommand{\ourmethod}{\texttt{\ourmethodname}}

\newcommand{\arenaname}{MemArena}
\newcommand{\arena}{\textsc{\arenaname}}
\newcommand{\pbenchname}{PersonaMind}
\newcommand{\pbench}{\pbenchname}

\newcommand{\Lverb}{\textsc{raw}}
\newcommand{\Lsum}{\textsc{summary}}
\newcommand{\Lgraph}{\textsc{graph}}
\newcommand{\Lskill}{\textsc{skill}}

\newcommand{\Lkv}{\textsc{kv memory}}

\newcommand{\lset}{\ensuremath{\mathcal{L}}}     
\newcommand{\vset}{\ensuremath{\mathcal{V}}}     
\newcommand{\iset}{\ensuremath{\mathcal{M}}}     
\newcommand{\supp}{\ensuremath{\mathrm{src}}}   
\newcommand{\srcof}{\ensuremath{\mathrm{in}}}    
\newcommand{\searchscope}{\ensuremath{F}}              
\newcommand{\qry}{\ensuremath{q}}                
\newcommand{\conf}{\ensuremath{c}}               
\newcommand{\ctx}{\ensuremath{C}}                
\newcommand{\itemx}{\ensuremath{x}}              
\newcommand{\itembest}{\ensuremath{x^{\star}}}   
\newcommand{\leaf}{\ensuremath{v}}               
\newcommand{\tokq}{\ensuremath{\mathrm{tok/q}}}  
\newcommand{\lrung}{\ensuremath{\ell}}          
\newcommand{\budget}{\ensuremath{B}}            

\newcommand{\snd}[1]{\second{#1}}   
\newcommand{\bestin}[2]{\textbf{#1}\,{\scriptsize\color{QOurs!85!black}(#2)}}   

\tcbset{claimbase/.style={enhanced, breakable, boxrule=0pt, arc=1.5pt, outer arc=1.5pt,
  left=6pt, right=6pt, top=4pt, bottom=4pt, before skip=4pt, after skip=4pt, fontupper=\normalsize}}
\newcounter{law}
\newtcolorbox{law}[1]{claimbase, colback=QOurs!5, colframe=QOurs!80!black,
  before upper={\refstepcounter{law}{\color{QOurs!80!black}\textbf{Observation~\thelaw\ (#1).}}\ }}
\newcounter{finding}
\newtcolorbox{finding}{claimbase, colback=C2!6, colframe=C2!85!black,
  before upper={\refstepcounter{finding}{\color{C2!85!black}\textbf{Finding~\thefinding.}}\ }}
\newtcolorbox{principle}{claimbase, colback=C3!7, colframe=C3!85!black,
  before upper={{\color{C3!85!black}\textbf{Design principle.}}\ }}

\newcommand{\chanTxt}{\ensuremath{\mathsf{C}_{\mathrm{txt}}}}   
\newcommand{\chanRt}{\ensuremath{\mathsf{C}_{\mathrm{route}}}}  

\newcommand{\accept}{\ensuremath{\mathsf{Accept}}}

\definecolor{QInk}{HTML}{1F2933}
\definecolor{QMute}{HTML}{8A94A6}
\definecolor{QFrame}{HTML}{5B6472}
\definecolor{QOurs}{HTML}{B4472F}
\colorlet{oursrow}{QOurs!9}
\definecolor{C0}{HTML}{B4472F}
\definecolor{C1}{HTML}{7C8794}
\definecolor{C2}{HTML}{2E6E8E}
\definecolor{C3}{HTML}{3E8E7E}
\definecolor{C4}{HTML}{8A6BAF}
\definecolor{C5}{HTML}{C08A2E}
\definecolor{C6}{HTML}{4A7CB5}
\definecolor{C7}{HTML}{6B8E3E}
\definecolor{C8}{HTML}{B5635A}

\newcommand{\rag}{\raggedright\let\\\tabularnewline}

\usepackage{listings}
\lstdefinestyle{api}{
  language=Python,
  basicstyle=\footnotesize\ttfamily\color{QInk},   
  keywordstyle=\color{C2}\bfseries,
  commentstyle=\color{QMute},
  stringstyle=\color{C3},
  emph={admit,index,score},            emphstyle=\color{QInk}\bfseries,
  emph={[2]route,propose},             emphstyle={[2]\color{QOurs}\bfseries},
  emph={[3]write,read,assemble},       emphstyle={[3]\color{C2}\bfseries},
  columns=fixed, basewidth={0.5em,0.45em}, keepspaces=true, showstringspaces=false,
  tabsize=4, breaklines=true, breakatwhitespace=true,
  frame=single, rulecolor=\color{black!35}, framerule=0.4pt, framesep=4pt,
  xleftmargin=0pt, xrightmargin=0pt, aboveskip=6pt, belowskip=4pt,
  escapeinside={(*@}{@*)},
  captionpos=b, abovecaptionskip=4pt, belowcaptionskip=0pt
}
\lstdefinestyle{trace}{
  style=api, language={}, morecomment=[l]{\#},
  emph={t1,t2,t3,t4,t5,t6,t7,t8,s1,s2,a1,a2,a3,a4,k1}, emphstyle=\color{C2}\bfseries,
  emph={[2]Stop,Narrow,Descend},       emphstyle={[2]\color{QOurs}\bfseries},
  emph={[3]},                          emphstyle={[3]}
}
\crefname{lstlisting}{Listing}{Listings}
\Crefname{lstlisting}{Listing}{Listings}
\definecolor{lossred}{HTML}{B03A2E}
\definecolor{gaingreen}{HTML}{2E7D4F}
\newcommand{\drop}[1]{{\color{lossred}\scriptsize(#1)}}
\newcommand{\rise}[1]{{\color{gaingreen}\scriptsize(#1)}}
\newcommand{\famrow}[2]{\multicolumn{#1}{l}{\footnotesize\textit{#2}}}

\algrenewcommand\algorithmicrequire{\textbf{Input:}\strut}
\algrenewcommand\algorithmicensure{\textbf{Output:}}
\AtBeginEnvironment{algorithmic}{\small}
\makeatletter
\expandafter\patchcmd\csname\string\algorithmic\endcsname
  {\labelwidth 1.2em}{\labelwidth 1.3em}{}{\PackageWarningNoLine{main}{algorithmic label patch failed}}
\makeatother
\DeclareRobustCommand{\Fn}[1]{\ensuremath{\mathop{\mathrm{#1}}\nolimits}}
\newcommand{\aStop}{\textsc{Stop}}
\newcommand{\aNarrow}{\textsc{Narrow}}
\newcommand{\aDescend}{\textsc{Descend}}
\algnewcommand{\IfThen}[2]{\State\algorithmicif\ #1\ \algorithmicthen\ #2}
\algnewcommand{\ElseThen}[1]{\State\algorithmicelse\ #1}
\tikzset{AlgHi/.style={draw=none,fill=McSoft,inner xsep=1pt,xshift=-1pt}}
\crefname{algorithm}{Algorithm}{Algorithms}
\Crefname{algorithm}{Algorithm}{Algorithms}
\crefname{ALG@line}{line}{lines}
\Crefname{ALG@line}{Line}{Lines}
\newcounter{algbody}
\makeatletter
\AtBeginEnvironment{algorithmic}{\stepcounter{algbody}\def\@currentcounter{ALG@line}%
  \ifdefined\algpx@indXLineLength\setlength{\algpx@indXLineLength}{0pt}\fi}
\def\theHALG@line{algbody.\arabic{algbody}.\arabic{ALG@line}}
\makeatother
\newlength{\AlgPairW}
\newsavebox{\AlgPairL}\newsavebox{\AlgPairR}\newlength{\AlgPairH}
\newcommand{\AlgPair}[6]{%
  \sbox{\AlgPairL}{\begin{minipage}[t]{\AlgPairW}\begin{algorithmic}[1]#3\end{algorithmic}\end{minipage}}%
  \sbox{\AlgPairR}{\begin{minipage}[t]{\AlgPairW}\begin{algorithmic}[1]#6\end{algorithmic}\end{minipage}}%
  \setlength{\AlgPairH}{\dimexpr\ht\AlgPairL+\dp\AlgPairL\relax}%
  \ifdim\dimexpr\ht\AlgPairR+\dp\AlgPairR\relax>\AlgPairH
    \setlength{\AlgPairH}{\dimexpr\ht\AlgPairR+\dp\AlgPairR\relax}\fi
  \noindent
  \begin{minipage}[t]{\AlgPairW}
    \begin{algorithm}[H]\captionsetup{hypcap=false}\caption[#1]{#1\strut}\phantomsection\label{#2}%
      \vbox to \AlgPairH{\copy\AlgPairL\vfil}%
    \end{algorithm}
  \end{minipage}\hfill
  \begin{minipage}[t]{\AlgPairW}
    \begin{algorithm}[H]\captionsetup{hypcap=false}\caption[#4]{#4\strut}\phantomsection\label{#5}%
      \vbox to \AlgPairH{\copy\AlgPairR\vfil}%
    \end{algorithm}
  \end{minipage}}

\makeatletter
\g@addto@macro\appendix{\crefalias{section}{appendix}\crefalias{subsection}{appendix}}
\makeatother
\crefname{appendix}{Appendix}{Appendices}
\Crefname{appendix}{Appendix}{Appendices}

\newcommand{\titleicon}{\raisebox{-2.4pt}{\includegraphics[height=13pt]{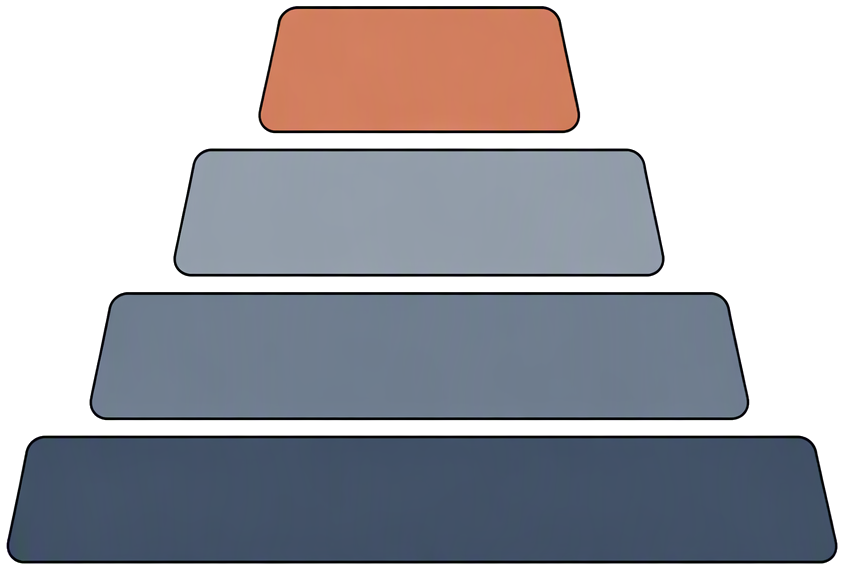}}\,}

%% file: figures/fig0_teaser.tex
\begin{tikzpicture}[font=\sffamily,
  slab/.style={text=QInk, font=\sffamily\fontsize{7}{8.4}\selectfont, inner sep=1.1pt},
  ourlab/.style={text=QInk, font=\sffamily\fontsize{7.4}{8.8}\bfseries\selectfont, inner sep=1.1pt},
  tk/.style={font=\sffamily\fontsize{7.2}{8.6}\selectfont, text=QInk},
  axl/.style={font=\sffamily\fontsize{8}{9.6}\bfseries\selectfont, text=QInk}]
\begin{axis}[at={(0cm,0cm)}, anchor=south west, scale only axis, width=6.3cm, height=3.5cm,
  xmin=0.5, xmax=4.5, ymin=0.5, ymax=3.5,
  xtick={1,2,3,4}, xticklabels={Flat,Indexed,Graph,Hierarchical},
  ytick={1,2,3}, yticklabels={{Hand-\\designed},{Fixed\\space},{Open-\\ended}},
  xticklabel style={tk}, yticklabel style={tk, rotate=90, anchor=south, yshift=1pt, align=center}, tick style={draw=none},
  axis line style={QFrame, line width=0.5pt}, xlabel={Memory organization}, ylabel={Design search},
  xlabel style={axl, yshift=1pt}, ylabel style={axl, yshift=-2pt},
  grid=both, grid style={QMute!30, line width=0.3pt, dash pattern=on 2pt off 1.6pt}, clip=false]
\fill[QOurs!14] (axis cs:3.5,2.5) rectangle (axis cs:4.5,3.5);
\addplot[only marks, mark=o, mark size=2.0pt, C1, mark options={line width=0.75pt}] coordinates {(1.00,1.00)};
\addplot[only marks, mark=+, mark size=2.4pt, C1, mark options={line width=0.75pt}] coordinates {(1.82,1.00)};
\addplot[only marks, mark=square, mark size=1.9pt, C2!70, mark options={line width=0.75pt}] coordinates {(2.18,1.00)};
\addplot[only marks, mark=triangle, mark size=2.3pt, C2!85, mark options={line width=0.75pt}] coordinates {(3.00,1.00)};
\addplot[only marks, mark=diamond, mark size=2.3pt, C2, mark options={line width=0.75pt}] coordinates {(3.78,1.00)};
\addplot[only marks, mark=pentagon, mark size=2.3pt, C2!55, mark options={line width=0.75pt}] coordinates {(4.22,1.00)};
\addplot[only marks, mark=otimes, mark size=2.3pt, C2!45, mark options={line width=0.75pt}] coordinates {(4.00,1.00)};
\addplot[only marks, mark=star, mark size=2.5pt, C5!85, mark options={line width=0.75pt}] coordinates {(2.00,2.00)};
\addplot[only marks, mark=asterisk, mark size=2.4pt, C5!70, mark options={line width=0.75pt}] coordinates {(3.00,2.00)};
\addplot[only marks, mark=halfcircle*, mark size=2.3pt, C5, mark options={line width=0.75pt}] coordinates {(4.00,2.00)};
\addplot[only marks, mark=oplus, mark size=2.3pt, C5!62, mark options={line width=0.75pt}] coordinates {(1.00,3.00)};
\addplot[only marks, mark=square*, mark size=3.0pt, mark options={fill=QOurs, draw=QInk, line width=0.7pt}] coordinates {(4.00,3.00)};
\node[slab, anchor=north] at (axis cs:1.00,0.90) {Full context};
\node[slab, anchor=south] at (axis cs:1.82,1.10) {Dense RAG};
\node[slab, anchor=north] at (axis cs:2.18,0.90) {Mem0};
\node[slab, anchor=north] at (axis cs:3.00,0.90) {A-Mem};
\node[slab, anchor=south] at (axis cs:3.66,1.10) {MemGPT};
\node[slab, anchor=south east] at (axis cs:4.49,1.30) {MemoryOS};
\node[slab, anchor=north] at (axis cs:4.00,0.90) {HMO};
\node[slab, anchor=south] at (axis cs:2.00,2.10) {AutoMem};
\node[slab, anchor=south] at (axis cs:3.00,2.10) {MemEvolve};
\node[slab, anchor=south] at (axis cs:4.00,2.10) {Mem-$\alpha$};
\node[slab, anchor=south] at (axis cs:1.00,3.10) {MemPro};
\node[ourlab, anchor=south] at (axis cs:4.00,3.12) {\ourmethod{}};
\draw[QMute, line width=0.4pt] (axis cs:4.22,1.06) -- (axis cs:4.22,1.28);
\end{axis}
\begin{axis}[at={(7.45cm,0cm)}, anchor=south west, scale only axis, width=5.0cm, height=3.5cm,
  xmin=1.8, xmax=10.6, ymin=46.5, ymax=62.5, xtick={2,4,6,8,10}, ytick={48,52,56,60},
  axis x line*=bottom, axis y line*=left, axis line style={QFrame, line width=0.5pt},
  xticklabel style={font=\sffamily\fontsize{7}{8.4}\selectfont, text=QMute, /pgf/number format/assume math mode=true},
  yticklabel style={font=\sffamily\fontsize{7}{8.4}\selectfont, text=QMute, /pgf/number format/assume math mode=true},
  xlabel={Latency per query (s)}, ylabel={Avg. accuracy (\%)},
  xlabel style={axl, yshift=1pt}, ylabel style={axl, yshift=-3pt},
  ymajorgrids=true, grid style={QMute!22, line width=0.25pt}, tick align=outside, clip=false]
\addplot[only marks, mark=o, mark size=2.0pt, C1, mark options={line width=0.75pt}] coordinates {(9.6,53.6)};
\addplot[only marks, mark=+, mark size=2.4pt, C1, mark options={line width=0.75pt}] coordinates {(3.1,47.6)};
\addplot[only marks, mark=square, mark size=1.9pt, C2!70, mark options={line width=0.75pt}] coordinates {(3.2,50.4)};
\addplot[only marks, mark=triangle, mark size=2.3pt, C2!85, mark options={line width=0.75pt}] coordinates {(3.9,49.3)};
\addplot[only marks, mark=diamond, mark size=2.3pt, C2, mark options={line width=0.75pt}] coordinates {(7.2,48.3)};
\addplot[only marks, mark=pentagon, mark size=2.3pt, C2!55, mark options={line width=0.75pt}] coordinates {(5.1,51.7)};
\addplot[only marks, mark=otimes, mark size=2.3pt, C2!45, mark options={line width=0.75pt}] coordinates {(7.6,52.5)};
\addplot[only marks, mark=star, mark size=2.5pt, C5!85, mark options={line width=0.75pt}] coordinates {(5.7,54.1)};
\addplot[only marks, mark=asterisk, mark size=2.4pt, C5!70, mark options={line width=0.75pt}] coordinates {(4.9,54.7)};
\addplot[only marks, mark=halfcircle*, mark size=2.3pt, C5, mark options={line width=0.75pt}] coordinates {(4.3,53.6)};
\addplot[only marks, mark=oplus, mark size=2.3pt, C5!62, mark options={line width=0.75pt}] coordinates {(5.6,55.2)};
\addplot[only marks, mark=square*, mark size=3.0pt, mark options={fill=QOurs, draw=QInk, line width=0.7pt}] coordinates {(2.7,60.8)};
\node[slab, anchor=south] at (axis cs:9.60,53.95) {Full context};
\node[slab, anchor=west] at (axis cs:3.32,47.60) {Dense RAG};
\node[slab, anchor=south] at (axis cs:3.20,50.70) {Mem0};
\node[slab, anchor=west] at (axis cs:4.12,49.30) {A-Mem};
\node[slab, anchor=west] at (axis cs:7.42,48.30) {MemGPT};
\node[slab, anchor=west] at (axis cs:5.32,51.70) {MemoryOS};
\node[slab, anchor=west] at (axis cs:7.82,52.50) {HMO};
\node[slab, anchor=west] at (axis cs:5.92,53.90) {AutoMem};
\node[slab, anchor=south east] at (axis cs:4.85,54.95) {MemEvolve};
\node[slab, anchor=east] at (axis cs:4.08,53.60) {Mem-$\alpha$};
\node[slab, anchor=south west] at (axis cs:5.65,55.45) {MemPro};
\node[ourlab, anchor=west] at (axis cs:2.95,60.80) {\ourmethod{}};
\end{axis}
\end{tikzpicture}

%% file: tables/main.tex
\begin{tabular}{l ccccc ccccc c}
\toprule
 & \multicolumn{5}{c}{Qwen3.5-27B} & \multicolumn{5}{c}{GPT-5.4-mini} & \\
\cmidrule(lr){2-6} \cmidrule(lr){7-11}
Method & LoCoMo & LME & MAB & PMind & Avg. & LoCoMo & LME & MAB & PMind & Avg. & Tok/q \\
\midrule
Full context & \snd{61.8} & 57.6 & 43.9 & 51.2 & 53.6 & \snd{64.1} & 59.8 & 46.2 & 53.3 & 55.9 & 21.4 \\
Dense RAG & 48.9 & 55.1 & 40.7 & 45.8 & 47.6 & 51.2 & 57.3 & 43.0 & 48.1 & 49.9 & \snd{3.6} \\
\midrule
MemGPT & 55.4 & 50.9 & 39.8 & 47.1 & 48.3 & 58.0 & 52.9 & 41.9 & 49.5 & 50.6 & 8.9 \\
A-Mem & 54.6 & 53.2 & 41.5 & 47.9 & 49.3 & 56.8 & 55.6 & 43.4 & 50.1 & 51.5 & 5.1 \\
Mem0 & 56.4 & 54.1 & 42.2 & 48.8 & 50.4 & 58.7 & 56.4 & 44.5 & 51.1 & 52.7 & 4.2 \\
MemoryOS & 58.0 & 54.8 & 43.6 & 50.3 & 51.7 & 60.3 & 57.0 & 45.9 & 52.6 & 54.0 & 6.3 \\
HMO & 58.7 & 55.9 & 44.1 & 51.3 & 52.5 & 60.9 & 58.3 & 46.2 & 53.4 & 54.7 & 7.4 \\
\midrule
Mem-$\alpha$ & 59.3 & 56.8 & 45.7 & 52.5 & 53.6 & 61.5 & 58.9 & 47.9 & 54.6 & 55.7 & 5.8 \\
AutoMem & 60.1 & 57.3 & 46.0 & 52.9 & 54.1 & 62.4 & 59.6 & 48.3 & 55.2 & 56.4 & 6.1 \\
MemEvolve & 60.6 & 58.0 & 46.6 & 53.5 & 54.7 & 62.9 & 60.3 & 48.9 & 55.8 & 57.0 & 6.6 \\
MemPro & 61.2 & \snd{58.4} & \snd{47.1} & \snd{54.0} & \snd{55.2} & 63.5 & \snd{60.7} & \snd{49.4} & \snd{56.2} & \snd{57.5} & 6.8 \\
\midrule
Initial $g_0$ & 63.4 & 59.7 & 48.3 & 55.7 & 56.8 & 65.3 & 61.6 & 50.1 & 57.1 & 58.5 & 2.4 \\
\rowcolor{oursrow}\ourmethod{} & \best{67.1} & \best{62.9} & \best{52.3} & \best{60.9} & \bestin{60.8}{$+$5.6} & \best{68.2} & \best{63.9} & \best{52.8} & \best{59.4} & \bestin{61.1}{$+$3.6} & \best{2.0} \\
\bottomrule
\end{tabular}

%% file: tables/ablation.tex
\begin{tabular}{@{}l ccc@{\hspace{3pt}}l ccc@{}}
\toprule
Method & Avg.\,(\%) & Rec.\,(\%) & Tok/q\,(k) & Method & Avg.\,(\%) & Rec.\,(\%) & Tok/q\,(k) \\
\midrule
\cellcolor{oursrow}\ourmethod{} & \cellcolor{oursrow}60.8 & \cellcolor{oursrow}88.4 & \cellcolor{oursrow}2.0 & \famrow{4}{Evolution} \\
\famrow{4}{Hierarchy depth} & \quad No evolution & 56.8\,\drop{$-$4.0} & 82.6\,\drop{$-$5.8} & 2.4 \\
\quad 1 layer (raw) & 54.1\,\drop{$-$6.7} & 71.2\,\drop{$-$17.2} & 1.8 & \quad Module selection & 58.3\,\drop{$-$2.5} & 84.9\,\drop{$-$3.5} & 2.3 \\
\quad 2 layers (+summary) & 56.5\,\drop{$-$4.3} & 78.9\,\drop{$-$9.5} & 2.3 & \quad Layer rewrites & 57.2\,\drop{$-$3.6} & 80.9\,\drop{$-$7.5} & 2.9 \\
\quad 3 layers (+graph) & 59.0\,\drop{$-$1.8} & 84.6\,\drop{$-$3.8} & 2.1 & \quad Program rewrites & 56.7\,\drop{$-$4.1} & 79.8\,\drop{$-$8.6} & 3.2 \\
\famrow{4}{Evidence channel} & \quad No acceptance test & 58.1\,\drop{$-$2.7} & 84.7\,\drop{$-$3.7} & 2.6 \\
\quad Content channel & 55.4\,\drop{$-$5.4} & 63.0\,\drop{$-$25.4} & 3.9 & \quad No archive & 59.2\,\drop{$-$1.6} & 86.8\,\drop{$-$1.6} & 2.1 \\
\quad Content + routing & 59.6\,\drop{$-$1.2} & 88.7\,\rise{$+$0.3} & 4.6 &  &  &  &  \\
\bottomrule
\end{tabular}

%% file: figures/fig4_dynamics.tex
\begin{tikzpicture}[font=\sffamily,
  tk/.style={font=\sffamily\fontsize{7}{8.4}\selectfont, text=QMute},
  lab/.style={font=\sffamily\fontsize{7}{8.2}\selectfont, text=QInk, inner sep=1pt},
  axl/.style={font=\sffamily\fontsize{8}{9.6}\bfseries\selectfont, text=QInk}]
\pgfplotsset{dyn/.style={scale only axis, axis x line*=bottom, axis y line*=left, axis line style={QFrame, line width=0.5pt},
  xticklabel style={tk, /pgf/number format/assume math mode=true}, yticklabel style={tk, /pgf/number format/assume math mode=true},
  xlabel style={axl, yshift=2pt}, ylabel style={axl, yshift=-3pt}, xtick pos=bottom, ytick pos=left,
  tick align=outside, tick style={QFrame, line width=0.4pt}, clip=false}}
\begin{axis}[dyn, at={(0cm,0cm)}, anchor=south west, width=2.8cm, height=2.55cm,
  xmin=-0.3, xmax=8.4, ymin=-0.25, ymax=4.15, xtick={0,2,4,6,8}, ytick={0,1,2,3,4},
  ymajorgrids=true, grid style={QMute!22, line width=0.25pt},
  xlabel={Evolution round}, ylabel={Gain over round 0 (points)},
  legend style={draw=none, fill=none, at={(0.5,1.02)}, anchor=south, legend columns=2, font=\sffamily\fontsize{6.8}{7.8}\selectfont, text=QInk,
    row sep=-2.6pt, column sep=3pt, inner sep=0.5pt}, legend cell align=left, legend image post style={xscale=0.75}]
\addplot[QOurs, solid, line width=0.9pt, mark=o, mark size=1.3pt, mark options={solid, fill=white, line width=0.6pt}] coordinates {(0,0.0) (1,1.8) (2,2.5) (3,2.7) (4,3.2) (5,3.4) (6,3.6) (7,3.6) (8,3.7)};
\addplot[QFrame, dashed, line width=0.7pt, mark=oplus, mark size=1.5pt, mark options={solid, fill=white, line width=0.6pt}] coordinates {(0,0.0) (1,0.2) (2,0.3) (3,0.9) (4,1.0) (5,1.7) (6,1.9) (7,2.4) (8,2.5)};
\addplot[QMute, dash dot, line width=0.7pt, mark=asterisk, mark size=1.6pt, mark options={solid, fill=white, line width=0.6pt}] coordinates {(0,0.0) (1,0.5) (2,0.9) (3,1.2) (4,1.4) (5,1.5) (6,1.7) (7,1.7) (8,1.7)};
\addplot[QMute!80, dotted, line width=0.75pt, mark=star, mark size=1.6pt, mark options={solid, fill=white, line width=0.6pt}] coordinates {(0,0.0) (1,0.9) (2,1.3) (3,1.5) (4,1.5) (5,1.6) (6,1.5) (7,1.6) (8,1.6)};
\legend{\ourmethod{},MemPro,MemEvolve,AutoMem}
\end{axis}
\begin{axis}[dyn, at={(5.0cm,0cm)}, anchor=south west, width=2.6cm, height=2.55cm,
  xbar stacked, bar width=5.2pt, xmin=0, xmax=100, xtick={0,50,100},
  ymin=-0.6, ymax=6.6, ytick={0,1,2,3,4,5,6}, yticklabels={{Single-hop},{Multi-hop},{Temporal},{Open-domain},{LME},{MAB},{PMind}}, y dir=reverse,
  yticklabel style={font=\sffamily\fontsize{7}{8.4}\selectfont, text=QInk}, ytick style={draw=none},
  xlabel={Reads stopping at layer (\%)}]
\addplot[draw=white, line width=0.3pt, fill=QFrame!85] coordinates {(3,0) (5,1) (2,2) (8,3) (0,4) (11,5) (16,6)};
\addplot[draw=white, line width=0.3pt, fill=QFrame!60] coordinates {(10,0) (29,1) (14,2) (31,3) (10,4) (24,5) (21,6)};
\addplot[draw=white, line width=0.3pt, fill=QFrame!36] coordinates {(22,0) (38,1) (41,2) (33,3) (23,4) (28,5) (35,6)};
\addplot[draw=white, line width=0.3pt, fill=QFrame!16] coordinates {(65,0) (28,1) (43,2) (28,3) (67,4) (37,5) (28,6)};
\draw[QMute, line width=0.3pt, dash pattern=on 1.5pt off 1pt] (axis cs:0,3.5) -- (axis cs:100,3.5);
\draw[QMute, line width=0.45pt] ([xshift=-1.72cm]axis cs:0,-0.3) -- ++(-2pt,0) |- ([xshift=-1.72cm]axis cs:0,3.3);
\node[lab, rotate=90, anchor=south] at ([xshift=-1.82cm]axis cs:0,1.5) {LoCoMo};
\fill[QFrame!85] (axis cs:-6,-2.30) rectangle ++(4.5pt,4.5pt); \node[lab, anchor=west, text height=1.5ex, text depth=0.3ex] at ([xshift=6pt,yshift=2.25pt]axis cs:-6,-2.30) {Skill};
\fill[QFrame!60] (axis cs:46,-2.30) rectangle ++(4.5pt,4.5pt); \node[lab, anchor=west, text height=1.5ex, text depth=0.3ex] at ([xshift=6pt,yshift=2.25pt]axis cs:46,-2.30) {Graph};
\fill[QFrame!36] (axis cs:-6,-1.58) rectangle ++(4.5pt,4.5pt); \node[lab, anchor=west, text height=1.5ex, text depth=0.3ex] at ([xshift=6pt,yshift=2.25pt]axis cs:-6,-1.58) {Summary};
\fill[QFrame!16] (axis cs:46,-1.58) rectangle ++(4.5pt,4.5pt); \node[lab, anchor=west, text height=1.5ex, text depth=0.3ex] at ([xshift=6pt,yshift=2.25pt]axis cs:46,-1.58) {Raw};
\end{axis}
\begin{axis}[dyn, at={(9.1cm,0cm)}, anchor=south west, width=3.6cm, height=2.55cm,
  ybar=0.7pt, bar width=4.4pt, xmin=-0.55, xmax=3.55, ymin=40, ymax=80, ytick={40,50,60,70,80},
  xtick={0,1,2,3}, xticklabels={{Single-hop},{Multi-hop},{Temporal},{Open-domain}}, xticklabel style={tk, text=QInk, rotate=18, anchor=north east, xshift=5pt, yshift=2.5pt},
  xtick style={draw=none}, ymajorgrids=true, grid style={QMute!22, line width=0.25pt},
  ylabel={LoCoMo test accuracy (\%)},
  legend style={draw=none, fill=none, at={(0.995,0.975)}, anchor=north east, legend columns=1, font=\sffamily\fontsize{6.8}{7.8}\selectfont, text=QInk,
    row sep=-3.2pt, inner sep=0.5pt}, legend cell align=left, legend image code/.code={\fill[#1] (0cm,-0.07cm) rectangle (0.2cm,0.1cm);}]
\addplot[draw=none, fill=QFrame!28] coordinates {(0,74.3) (1,49.2) (2,56.0) (3,47.4)};
\addplot[draw=none, fill=QFrame!62] coordinates {(0,67.3) (1,52.8) (2,57.2) (3,50.6)};
\addplot[draw=none, fill=QOurs] coordinates {(0,74.6) (1,57.4) (2,60.1) (3,52.8)};
\legend{Always raw (3.3k),Stop at first hit (1.6k),Adaptive (1.9k)}
\end{axis}
\end{tikzpicture}

%% file: tables/chain.tex
\begin{tabular}{c l ccc}
\toprule
Round & Edit & Evolve & Test & Tok/q\,(k) \\
\midrule
0 & $g_0$ & 63.6 & 63.4 & 2.31 \\
1 & \Lsum{}.\texttt{route} & 66.9 & 65.2 & 2.26 \\
2 & \Lgraph{}.\texttt{admit} & 68.2 & 65.9 & 2.21 \\
3 & \Lverb{}.\texttt{score} & 68.5 & 66.1 & 2.13 \\
5 & \Lsum{}.\texttt{index} & 69.5 & 66.8 & 2.02 \\
7 & \Lskill{}.\texttt{route} & 69.9 & 67.0 & 1.94 \\
\rowcolor{oursrow}8 & \Lgraph{}.\texttt{propose} & 70.1 & 67.1 & 1.93 \\
\bottomrule
\end{tabular}

%% file: tables/record_split.tex
\begin{tabular}{lccccc}
\toprule
Method & LoCoMo & LongMemEval & MAB & PersonaMind & Avg. \\
\midrule
AutoMem & 59.6 & 57.3 & 45.5 & 52.2 & 53.7 {\scriptsize($-$0.4)} \\
MemEvolve & 60.0 & 58.0 & 46.1 & 52.9 & 54.3 {\scriptsize($-$0.4)} \\
MemPro & 60.7 & 58.4 & 46.6 & 53.5 & 54.8 {\scriptsize($-$0.4)} \\
\ourmethod{} (initial $g_0$) & \snd{63.0} & \snd{59.7} & \snd{47.9} & \snd{55.1} & 56.4 {\scriptsize($-$0.4)} \\
\rowcolor{oursrow}\ourmethod{} & \best{65.6} & \best{62.9} & \best{51.4} & \best{59.3} & \best{59.8} {\scriptsize($-$1.0)} \\
\bottomrule
\end{tabular}

%% file: figures/fig6_results.tex
\begin{tikzpicture}[font=\sffamily,
  tk/.style={font=\sffamily\fontsize{7}{8.4}\selectfont, text=QMute},
  lab/.style={font=\sffamily\fontsize{7}{8.2}\selectfont, text=QInk, inner sep=1pt},
  axl/.style={font=\sffamily\fontsize{8}{9.6}\bfseries\selectfont, text=QInk}]
\pgfplotsset{dyn/.style={scale only axis, axis x line*=bottom, axis y line*=left, axis line style={QFrame, line width=0.5pt},
  xticklabel style={tk, /pgf/number format/assume math mode=true}, yticklabel style={tk, /pgf/number format/assume math mode=true},
  xlabel style={axl, yshift=2pt}, ylabel style={axl, yshift=-3pt}, xtick pos=bottom, ytick pos=left,
  tick align=outside, tick style={QFrame, line width=0.4pt}, clip=false}}
\begin{axis}[dyn, at={(0.00cm,0cm)}, anchor=south west, width=5.55cm, height=2.55cm,
  ybar, bar width=3.9pt, xmin=-0.6, xmax=12.6, ymin=50, ymax=100, ytick={50,60,70,80,90,100},
  xtick={0,1,2,3,4,5,6,7,8,9,10,11,12}, xticklabels={{Full context},{Dense RAG},{MemGPT},{A-Mem},{Mem0},{MemoryOS},{HMO},{Mem-$\alpha$},{AutoMem},{MemEvolve},{MemPro},{Initial $g_0$},{\ourmethod{}}},
  xticklabel style={tk, text=QInk, rotate=42, anchor=north east, xshift=3.5pt, yshift=1pt}, xtick style={draw=none},
  ymajorgrids=true, grid style={QMute!22, line width=0.25pt}, ylabel={Evidence recall (\%)},
  legend style={draw=none, fill=none, at={(0.02,1.0)}, anchor=south west, legend columns=2, font=\sffamily\fontsize{6.8}{7.8}\selectfont, text=QInk,
    column sep=3pt, inner sep=0.5pt}, legend cell align=left, legend image code/.code={\fill[#1] (0cm,-0.07cm) rectangle (0.2cm,0.1cm);}]
\addplot[draw=none, fill=QFrame!62, bar shift=-2.75pt] coordinates {(0,97.8) (1,74.6) (2,62.5) (3,66.0) (4,68.7) (5,72.9) (6,74.1) (7,75.8) (8,77.2) (9,78.0) (11,82.6)};
\addplot[draw=none, fill=QFrame!28, bar shift=2.75pt] coordinates {(0,64.5) (1,88.4) (2,70.3) (3,79.8) (4,81.5) (5,83.0) (6,84.6) (7,86.1) (8,87.0) (9,88.2) (11,90.3)};
\legend{LoCoMo,LongMemEval}
\addplot[draw=none, fill=QOurs, bar shift=-2.75pt, forget plot] coordinates {(12,88.4)};
\addplot[draw=none, fill=QOurs!50, bar shift=2.75pt, forget plot] coordinates {(12,93.7)};
\node[font=\sffamily\fontsize{6}{6.6}\selectfont, text=QInk, rotate=90, anchor=west, inner xsep=1.2pt, inner ysep=0pt] at ([xshift=-2.75pt]axis cs:12,88.4) {88.4};
\node[font=\sffamily\fontsize{6}{6.6}\selectfont, text=QInk, rotate=90, anchor=west, inner xsep=1.2pt, inner ysep=0pt] at ([xshift=2.75pt]axis cs:12,93.7) {93.7};
\addplot[draw=none, fill=QFrame!62, bar shift=-2.75pt, forget plot] coordinates {(10,79.3)};
\addplot[draw=none, fill=QFrame!28, bar shift=2.75pt, forget plot] coordinates {(10,89.1)};
\node[font=\sffamily\fontsize{6}{6.6}\selectfont, text=QFrame, rotate=90, anchor=west, inner xsep=1.2pt, inner ysep=0pt] at ([xshift=-2.75pt]axis cs:10,79.3) {79.3};
\node[font=\sffamily\fontsize{6}{6.6}\selectfont, text=QFrame, rotate=90, anchor=west, inner xsep=1.2pt, inner ysep=0pt] at ([xshift=2.75pt]axis cs:10,89.1) {89.1};
\end{axis}
\begin{axis}[dyn, at={(7.05cm,0cm)}, anchor=south west, width=3.40cm, height=2.55cm,
  ybar=0.7pt, bar width=4.2pt, xmin=-0.55, xmax=3.55, ymin=40, ymax=80, ytick={40,50,60,70,80},
  xtick={0,1,2,3}, xticklabels={{Single-hop},{Multi-hop},{Temporal},{Open-domain}}, xticklabel style={tk, text=QInk, rotate=18, anchor=north east, xshift=5pt, yshift=2.5pt},
  xtick style={draw=none}, ymajorgrids=true, grid style={QMute!22, line width=0.25pt},
  ylabel={LoCoMo test accuracy (\%)},
  legend style={draw=none, fill=none, at={(0.995,0.975)}, anchor=north east, legend columns=1, font=\sffamily\fontsize{6.8}{7.8}\selectfont, text=QInk,
    row sep=-3.2pt, inner sep=0.5pt}, legend cell align=left, legend image code/.code={\fill[#1] (0cm,-0.07cm) rectangle (0.2cm,0.1cm);}]
\addplot[draw=none, fill=QMute] coordinates {(0,70.5) (1,49.6) (2,52.0) (3,44.8)};
\addplot[draw=none, fill=QOurs!40] coordinates {(0,72.1) (1,51.8) (2,55.2) (3,47.9)};
\addplot[draw=none, fill=QOurs] coordinates {(0,74.6) (1,57.4) (2,60.1) (3,52.8)};
\legend{MemPro,Initial $g_0$,\ourmethod{}}
\end{axis}
\fill[QOurs!55] (12.150,1.912) rectangle ++(0.730,0.607); \node[font=\sffamily\fontsize{7}{8}\bfseries\selectfont, text=white] at (12.515,2.216) {67.1};
\fill[QOurs!23] (12.910,1.912) rectangle ++(0.730,0.607); \node[font=\sffamily\fontsize{7}{8}\selectfont, text=QInk] at (13.275,2.216) {61.0};
\fill[QOurs!24] (13.670,1.912) rectangle ++(0.730,0.607); \node[font=\sffamily\fontsize{7}{8}\selectfont, text=QInk] at (14.035,2.216) {49.6};
\fill[QOurs!33] (14.430,1.912) rectangle ++(0.730,0.607); \node[font=\sffamily\fontsize{7}{8}\selectfont, text=QInk] at (14.795,2.216) {57.8};
\node[lab, anchor=east] at (12.110,2.216) {LoCoMo};
\fill[QOurs!26] (12.150,1.275) rectangle ++(0.730,0.607); \node[font=\sffamily\fontsize{7}{8}\selectfont, text=QInk] at (12.515,1.579) {64.9};
\fill[QOurs!55] (12.910,1.275) rectangle ++(0.730,0.607); \node[font=\sffamily\fontsize{7}{8}\bfseries\selectfont, text=white] at (13.275,1.579) {62.9};
\fill[QOurs!15] (13.670,1.275) rectangle ++(0.730,0.607); \node[font=\sffamily\fontsize{7}{8}\selectfont, text=QInk] at (14.035,1.579) {48.9};
\fill[QOurs!19] (14.430,1.275) rectangle ++(0.730,0.607); \node[font=\sffamily\fontsize{7}{8}\selectfont, text=QInk] at (14.795,1.579) {56.6};
\node[lab, anchor=east] at (12.110,1.579) {LME};
\fill[QOurs!18] (12.150,0.637) rectangle ++(0.730,0.607); \node[font=\sffamily\fontsize{7}{8}\selectfont, text=QInk] at (12.515,0.941) {64.2};
\fill[QOurs!15] (12.910,0.637) rectangle ++(0.730,0.607); \node[font=\sffamily\fontsize{7}{8}\selectfont, text=QInk] at (13.275,0.941) {60.3};
\fill[QOurs!56] (13.670,0.637) rectangle ++(0.730,0.607); \node[font=\sffamily\fontsize{7}{8}\bfseries\selectfont, text=white] at (14.035,0.941) {52.3};
\fill[QOurs!37] (14.430,0.637) rectangle ++(0.730,0.607); \node[font=\sffamily\fontsize{7}{8}\selectfont, text=QInk] at (14.795,0.941) {58.1};
\node[lab, anchor=east] at (12.110,0.941) {MAB};
\fill[QOurs!34] (12.150,0.000) rectangle ++(0.730,0.607); \node[font=\sffamily\fontsize{7}{8}\selectfont, text=QInk] at (12.515,0.304) {65.6};
\fill[QOurs!26] (12.910,0.000) rectangle ++(0.730,0.607); \node[font=\sffamily\fontsize{7}{8}\selectfont, text=QInk] at (13.275,0.304) {61.2};
\fill[QOurs!33] (13.670,0.000) rectangle ++(0.730,0.607); \node[font=\sffamily\fontsize{7}{8}\selectfont, text=QInk] at (14.035,0.304) {50.4};
\fill[QOurs!70] (14.430,0.000) rectangle ++(0.730,0.607); \node[font=\sffamily\fontsize{7}{8}\bfseries\selectfont, text=white] at (14.795,0.304) {60.9};
\node[lab, anchor=east] at (12.110,0.304) {PMind};
\node[lab, anchor=north] at (12.515,-0.02) {LoCoMo};
\node[lab, anchor=north] at (13.275,-0.02) {LME};
\node[lab, anchor=north] at (14.035,-0.02) {MAB};
\node[lab, anchor=north] at (14.795,-0.02) {PMind};
\node[axl, anchor=north] at (13.655,-0.34) {Evaluated on};
\node[axl, anchor=south, rotate=90] at (11.230,1.260) {Evolved on};
\end{tikzpicture}